\documentclass{article}

\usepackage{graphicx} 
\usepackage{subfigure}
\usepackage{wrapfig}
\usepackage{xspace}
\usepackage{caption}
\usepackage{booktabs}

\usepackage{natbib}

\usepackage{algorithm}
\usepackage{algorithmic}

\usepackage{enumitem}
\usepackage{hyperref}

\usepackage{fullpage}
\usepackage{authblk}

\usepackage{Definitions}
\usepackage{color}

\graphicspath{{./Figs/}}

\renewcommand{\geq}{\geqslant}

\title{\huge Discriminative Embeddings of Latent Variable Models for Structured Data}

\author{
    Hanjun Dai, Bo Dai, Le Song\\
    College of Computing, Georgia Institute of Technology\\
    \{hanjundai, bodai\}@gatech.edu, lsong@cc.gatech.edu\\
}
\date{}

\begin{document}
\maketitle

\begin{abstract}
	Kernel classifiers and regressors designed for structured data, such as sequences, trees and graphs, have significantly advanced a number of interdisciplinary areas such as computational biology and drug design. Typically, kernels are designed beforehand for a data type which either exploit statistics of the structures or make use of probabilistic generative models, and then a discriminative classifier is learned based on the kernels via convex optimization. However, such an elegant two-stage approach also limited kernel methods from scaling up to millions of data points, and exploiting discriminative information to learn feature representations. 

	We propose, {\tt structure2vec}, an effective and scalable approach for structured data representation based on the idea of embedding latent variable models into feature spaces, and learning such feature spaces using discriminative information. Interestingly, {\tt structure2vec} extracts features by performing a sequence of function mappings in a way similar to graphical model inference procedures, such as mean field and belief propagation. In applications involving millions of data points, we showed that {\tt structure2vec} runs 2 times faster, produces models which are $10,000$ times smaller, while at the same time achieving the state-of-the-art predictive performance.
\end{abstract}

\section{Introduction}\label{sec:intro}

\setlength{\abovedisplayskip}{3pt}
\setlength{\abovedisplayshortskip}{1pt}
\setlength{\belowdisplayskip}{3pt}
\setlength{\belowdisplayshortskip}{1pt}
\setlength{\jot}{2pt}

\setlength{\floatsep}{2ex}
\setlength{\textfloatsep}{2ex}

Structured data, such as sequences, trees and graphs, are prevalent in a number of interdisciplinary areas such as protein design, genomic sequence analysis, and drug design~\citep{SchTsuVer04}. To learn from such complex data, we have to first transform such data explicitly or implicitly into some vectorial representations, and then apply machine learning algorithms in the resulting vector space. So far kernel methods have emerged as one of the most effective tools for dealing with structured data, and have achieved the state-of-the-art classification and regression results in many sequence~\citep{LesEskNob02,VisSmo03} and graph datasets~\citep{GaeFlaWro03,Borgwardt07}. 

The success of kernel methods on structured data relies crucially on the design of kernel functions --- positive semidefinite similarity measures between pairs of data points~\citep{SchSmo02}. By designing a kernel function, we have implicitly chosen a corresponding feature representation for each data point which can potentially has infinite dimensions. Later learning algorithms for various tasks and with potentially very different nature can then work exclusively on these pairwise kernel values without the need to access the original data points. Such modular structure of kernel methods has been very powerful, making them the most elegant and convenient methods to deal with structured data. Thus designing kernel for different structured objects, such as strings, trees and graphs, has always been an important subject in the kernel community. However, in the big data era, this modular framework has also limited kernel methods in terms of their ability to scale up to millions of data points, and exploit discriminative information to learn feature representations. 

For instance, a class of kernels are designed based on the idea of ``bag of structures'' (BOS), where each structured data point is represented as a vector of counts for elementary structures. The spectrum kernel and variants for strings~\citep{LesEskNob02}, subtree kernel~\citep{RamGae03}, graphlet kernel~\citep{SheVisPetMehetal09} and Weisfeiler-lehman graph kernel~\citep{SheSchVanMehetal11} all follow this design principle. In other words, the feature representations of these kernels are fixed before learning, with each dimension corresponding to a substructure, independent of the supervised learning tasks at hand. Since there are many unique substructures which may or may not be useful for the learning tasks, the explicit feature space of such kernels typically has very high dimensions. Subsequently algorithms dealing with the pairwise kernel values have to work with a big kernel matrix squared in the number of data points. The square dependency on the number of data points largely limits these BOS 
kernels to datasets of size just thousands. 

A second class of kernels are based on the ingenious idea of exploiting the ability of probabilistic graphical models (GM) in describing noisy and structured data to design kernels. For instance, one can use hidden Markov models for sequence data, and use pairwise Markov random fields for graph data. The Fisher kernel~\citep{JaaHau99b} and probability product kernel~\citep{JebKonHow04} are two representative instances within the family. The former method first fits a common generative model to the entire dataset, and then uses the empirical Fisher information matrix and the Fisher score of each data point to define the kernel; The latter method instead fits a different generative model for each data point, and then uses inner products between distributions to define the kernel. Typically the parameterization of these GM kernels are chosen before hand. Although the process of fitting generative models allow the kernels to adapt to the geometry of the input data, the resulting feature representations 
are still independent of the discriminative task at hand. Furthermore, the extra step of fitting generative models to data can be a challenging computation and estimation task by itself, especially in the presence of latent variables. Very often in practice, one finds that BOS kernels are easier to deploy than GM kernels, although the latter is supposed to capture the additional geometry and uncertainty information of data. 

In this paper, we wish to revisit the idea of using graphical models for kernel or feature space design, with the goal of scaling up kernel methods for structured data to millions of data points, and allowing the kernel to learn the feature representation from label information. Our idea is to model each structured data point as a latent variable model, then embed the graphical model into feature spaces~\citep{SmoGreSonSch07,SonHuaSmoFuk09}, and use inner product in the embedding space to define kernels. Instead of fixing a feature or embedding space beforehand, we will also learn the feature space by directly minimizing the empirical loss defined by the label information. 

The resulting embedding algorithm, {\tt structure2vec}, runs in a scheme similar to graphical model inference procedures, such as mean field and belief propagation. Instead of performing probabilistic operations (such as sum, product and renormalization), the algorithm performs nonlinear function mappings in each step, inspired by kernel message passing algorithm in~\citet{SonGreGue10, SonGreBicLowGue11}. Furthermore, {\tt structure2vec} is also different from the kernel message passing algorithm in several aspects. First, {\tt structure2vec} deals with a different scenario, \ie, learning similarity measure for structured data. Second, {\tt structure2vec} learns the nonlinear mappings using the discriminative information. And third, a variant of {\tt structure2vec} can run in a mean field update fashion, different from message passing algorithms. 

Besides the above novel aspects, {\tt structure2vec} is also very scalable in terms of both memory and computation requirements. First, it uses a small and explicit feature map for the nonlinear feature space, and avoids the need for keeping the kernel matrix. This makes the subsequent classifiers or regressors order of magnitude smaller compared to other methods. Second, the nonlinear function mapping in {\tt structure2vec} can be learned using stochastic gradient descent, allowing it to handle extremely large scale datasets.  

Finally in experiments, we show that {\tt structure2vec} compares favorably to other kernel methods in terms of classification accuracy in medium scale sequence and graph benchmark datasets including SCOP and NCI. Furthermore, {\tt structure2vec} can handle extremely large data set, such as the 2.3 million molecule dataset from Harvard Clean Energy Project, run 2 times faster, produce model $10,000$ times smaller and achieve state-of-the-art accuracy. These strong empirical results suggest that the graphical models, theoretically well-grounded methods for capturing structure in data, combined with embedding techniques and discriminative training can significantly improve the performance in many large scale real-world structured data classification and regression problems. 
  
\section{Backgrounds}\label{sec:background}

We denote by $X$ a random variable with domain $\Xcal$, and refer to instantiations of $X$ by the lower case character, $x$. We denote a density on $\Xcal$ by $p(X)$, and denote the space of all such densities by $\Pcal$. We will also deal with multiple random variables, $X_1, X_2, \ldots, X_{\ell}$, with joint density $p(X_1,X_2,\ldots,X_{\ell})$. For simplicity of notation, we assume that the domains of all $X_t, t \in [\ell]$ are the same, but the methodology applies to the cases where they have different domains. In the case when $\Xcal$ is a discrete domain, the density notation should be interpreted as probability, and integral should be interpreted as summation instead. Furthermore, we denote by $H$ a hidden variable with domain $\Hcal$ and distribution $p(H)$. We use similar notation convention for variable $H$ and $X$. \\ [-2mm]

\noindent
{\bf Kernel Methods.} Suppose the structured data is represented by $\chi \in \Gcal$.  Kernel methods owe the name to the use of kernel functions, $k(\chi,\chi'):\Gcal \times \Gcal \mapsto \RR$, which are symmetric positive semidefinite (PSD), meaning that for all $n > 1$, and $\chi_1,\ldots,\chi_n \in \Gcal$, and $c_1,\ldots,c_n \in \RR$, we have $\sum_{i,j=1}^n c_i c_j k(\chi_i, \chi_j) \geqslant 0$. A signature of kernel methods is that learning algorithms for various tasks and with potentially very different nature can work exclusively on these pairwise kernel values without the need to access the original data points. \\[-2mm]

\noindent
{\bf Kernels for Structured Data.} Each kernel function will correspond to some feature map $\phi(\chi)$, where the kernel function can be expressed as the inner product between feature maps,~\ie, $k(\chi,\chi')=\inner{\phi(\chi)}{\phi(\chi')}$. For structured input domain, one can design kernels using counts on substructures. For instance, the spectrum kernel for two sequences $\chi$ and $\chi'$ is defined as~\citep{LesEskNob02}
\begin{align}\label{eq:bos_kernel}
  k(\chi,\chi') = \sum\nolimits_{s \in \Scal} \#(s \in \chi) \#(s \in \chi')
\end{align}
where $\Scal$ is the set of possible subsequences, $\#(s \in x)$ counts the number occurrence of subsequence $s$ in $x$. In this case, the feature map $\phi(\chi) = (\#(s_1 \in \chi), \#(s_2 \in \chi), ...)^\top$ corresponds to a vector of dimension $\abr{\Scal}$. Similarly, the graphlet kernel~\citep{SheVisPetMehetal09} for two graphs $\chi$ and $\chi'$ can also be defined as~\eq{eq:bos_kernel}, but $\Scal$ is now the set of possible subgraphs, and $\#(s \in \chi)$ counts the number occurrence of subgraphs. We refer to this class of kernels as ``bag of structures'' (BOS) kernel. 

Kernels can also be defined by leveraging the power of probabilistic graphical models. For instance, the Fisher kernel~\citep{JaaHau99b} is defined using a parametric model $p(\chi|\theta^*)$ around its maximum likelihood estimate $\theta^*$,~\ie, 
$
k(\chi, \chi')=U_{\chi}^\top I^{-1} U_{\chi'}, 
$
where $U_{\chi} := \nabla_{\theta=\theta^*} \log p(\chi|\theta)$ and $I=\EE_{\Gcal}[U_{\Gcal} U_{\Gcal}^\top]$ is the Fisher information matrix. Another classical example along the line is the probability product kernel~\citep{JebKonHow04}. Different from the Fisher kernel based on generative model fitted with the whole dataset, the probability product kernel is calculated based on the models $p(\chi|\theta)$ fitted to individual data point, \ie, $k(\chi,\chi')= \int_{\Gcal} p(\tau|\theta_{\chi})^\rho p(\tau|\theta_{\chi'})^\rho d\tau$ where $\theta_{\chi}$ and $\theta_{\chi'}$ are the maximum likelihood parameters for data point $\chi$ and $\chi'$ respectively. We refer to this class of kernels as the ``graphical model'' (GM) kernels. \\ [-2mm]

\noindent
{\bf Hilbert Space Embedding of Distributions.} Hilbert space embeddings of distributions are mappings of distributions into potentially \emph{infinite} dimensional feature spaces~\citep{SmoGreSonSch07},
\begin{align}
  \mu_{X} \, := \, \EE_{X} \sbr{\phi(X)} \, = \, \int_{\Xcal} \phi(x) p(x) dx~:~ \Pcal \mapsto \Fcal \label{eq:embedding}
\end{align}
where the distribution is mapped to its expected feature map,~\ie,~to a point in a feature space. Kernel embedding of distributions has rich representational power. Some feature map can make the mapping injective~\citep{SriGreFukLanetal08}, meaning that if two distributions, $p(X)$ and $q(X)$, are different, they are mapped to two distinct points in the feature space. For instance, when $\Xcal=\RR^d$, the feature spaces of many commonly used kernels, such as the Gaussian RBF kernel $\exp(-\nbr{x-x'}_2^2)$, can make the embedding injective. 

Alternatively, one can treat an injective embedding $\mu_X$ of a density $p(X)$ as a sufficient statistic of the density. Any information we need from the density is preserved in $\mu_X$: with $\mu_X$ one can uniquely recover $p(X)$, and any operation on $p(X)$ can be carried out via a corresponding operation on $\mu_X$ with the same result. For instance, this property will allow us to compute a functional $f:\Pcal \mapsto \RR$ of the density using the embedding only,~\ie,
\begin{align}
  f(p(x)) = \ftil (\mu_X)  \label{eq:embed_funtional}
\end{align}
where $\ftil:\Fcal\mapsto \RR$ is a corresponding function applied on $\mu_X$. Similarly the property can also be generalized to operators. For instance, applying an operator $\Tcal:\Pcal \mapsto \RR^d$ to a density can also be equivalently carried out using its embedding,~\ie,
\vspace{-2mm}
\begin{align}
  \Tcal \circ p(x) &= \widetilde \Tcal \circ \mu_X,  \label{eq:embed_operator} 
\end{align}
where $\widetilde \Tcal: \Fcal\mapsto \RR^d$ is the alternative operator working on the embedding. In our later sections, we will extensively exploit this property of injective embeddings, by assuming that there exists a feature space such that the embeddings are injective. We include the discussion of other related work in Appendix~\ref{sec:more_related_work}. 

\section{Model for a Structured Data Point}\label{sec:model}

\begin{figure}[t] 
\small 
\centering 
\subfigure[b][Represent string data as a latent variable model ]
{
\includegraphics[width=0.48\linewidth]{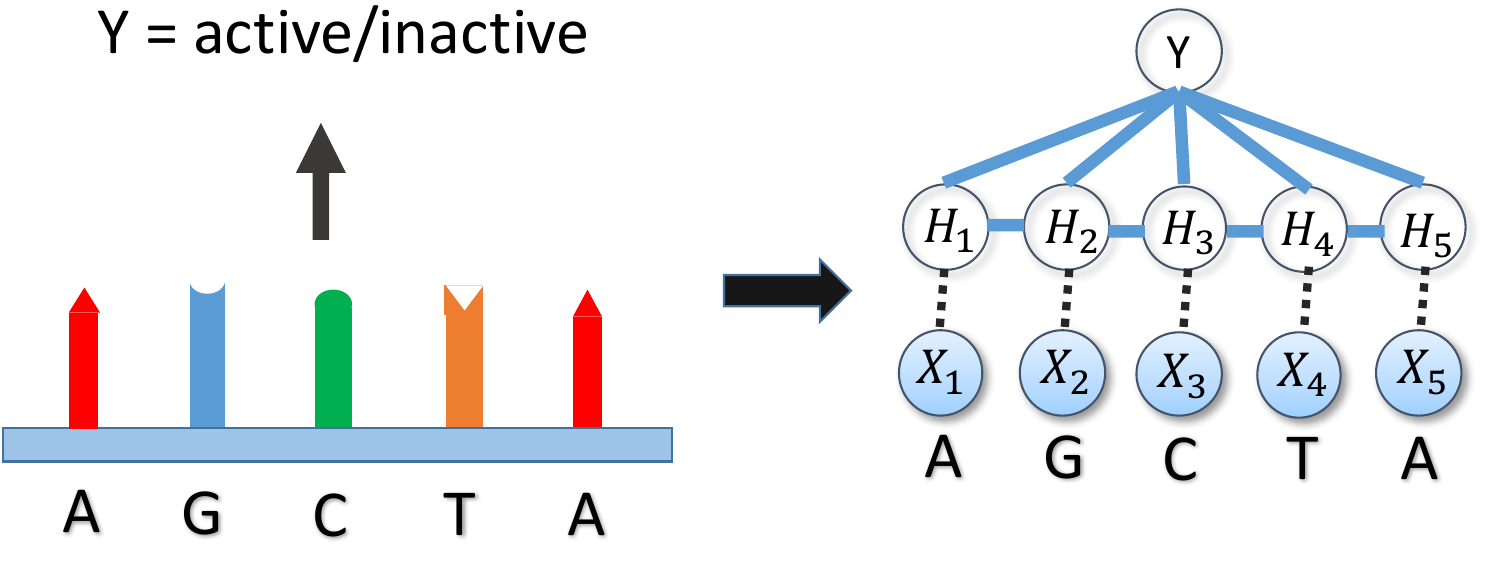} 
}
\hspace{3mm}
\subfigure[b][Represent graph data as a latent variable model]
{
\includegraphics[width=0.46\linewidth]{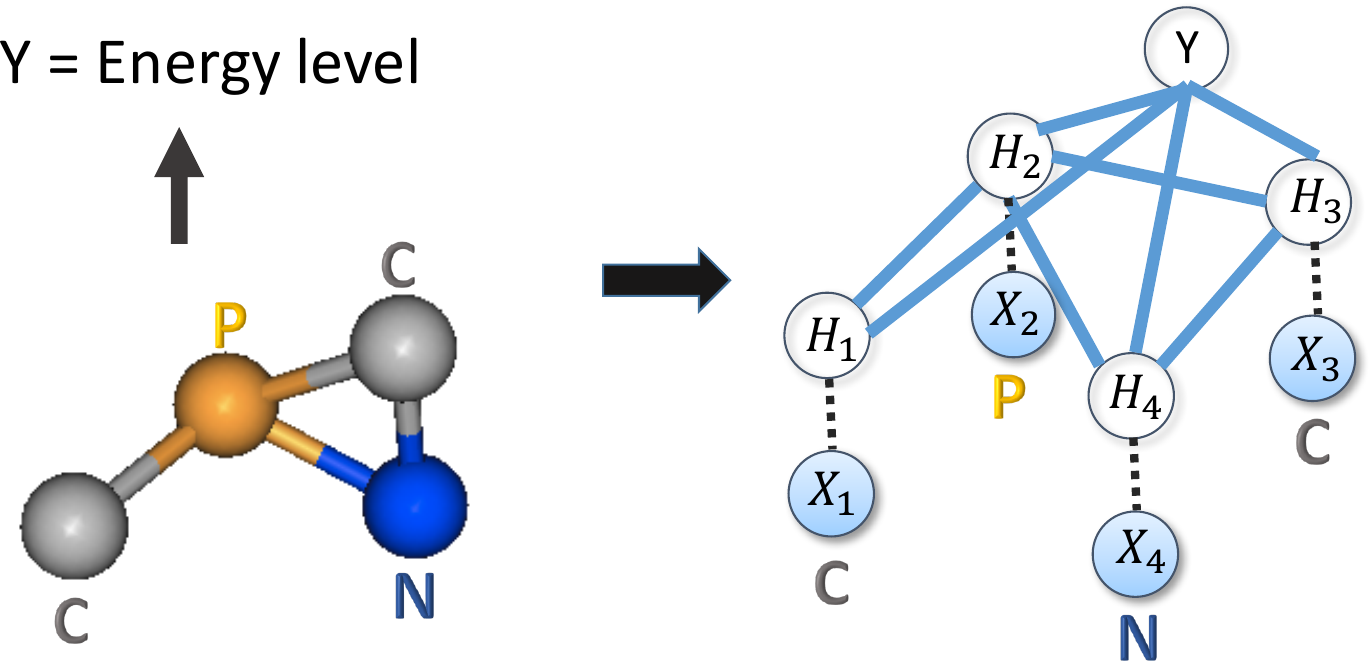} 
}
 \caption{ Building graphical model with hidden variables from structured string and general graph data. $Y$ is the supervised information, which can be real number (for regression) or discrete integer (for classification). }
 \label{fig:diagram_mf_lbp}
 \end{figure}

Without loss of generality, we assume each structured data point $\chi$ is a graph, with a set of nodes $\Vcal = \{1,\ldots, V\}$ and a set of edges $\Ecal$. We will use $x_i$ to denote the value of the attribute for node $i$. We note the node attributes are different from the label of the entire data point. For instance, each atom in a molecule will correspond to a node in the graph, and the node attribute will be the atomic number, while the label for the entire molecule can be whether the molecule is a good drug or not. Other structures, such as sequences and trees, can be viewed as special cases of general graphs. 

We will model the structured data point $\chi$ as an instance drawn from a graphical model. More specifically, we will model the label of each node in the graph with a variable $X_i$, and furthermore, associate an additional hidden variable $H_i$ with it. Then we will define a pairwise Markov random field on these collection of random variables
\begin{align}\label{eq:data_model}
  p(\cbr{H_i},\cbr{X_i}) \propto \prod_{i \in \Vcal} \Phi(H_i, X_i) \prod_{(i,j) \in \Ecal} \Psi(H_i,H_j)  
\end{align}
where $\Psi$ and $\Phi$ are nonnegative node and edge potentials respectively. In this model, the variables are connected according to the graph structure of the input data point. That is to say, we use the graph structure of the input data directly as the conditional independence structure of an undirected graphical model. Figure~\ref{fig:diagram_mf_lbp} illustrates two concrete examples in constructing the graphical models for strings and graphs. One can design more complicated graphical models which go beyond pairwise Markov random fields, and consider longer range interactions with potentials involving more variables. We will focus on pairwise Markov random fields for simplicity of representation.

We note that such a graphical model is built for each individual data point, and the conditional independence structures of two graphical models can be different if the two data points $\chi$ and $\chi'$ are different. Furthermore, we do not observe the value for the hidden variables $\cbr{H_i}$, which makes the learning of the graphical model potentials $\Phi$ and $\Psi$ even more difficult. Thus, we will not pursue the standard route of maximum likelihood estimation, and rather we will consider the sequence of computations needed when we try to embed the posterior of $\cbr{H_i}$ into a feature space.  

\section{Embedding Latent Variable Models}\label{sec:lvm_embedding}

We will embed the posterior marginal $p(H_i | \cbr{x_i})$ of a hidden variable using a feature map $\phi(H_i)$,~\ie, 
\begin{align}
  \mu_i = \int_{\Hcal} \phi(h_i) p(h_i | \cbr{x_i}) dh_i. 
\end{align}
The exact form of $\phi(H_i)$ and the parameters in MRF $p(H_i | \cbr{x_i})$ is not fixed at the moment, and we will learn them later using supervision signals for the ultimate discriminative target. For now, we will assume that $\phi(H_i) \in \RR^d$ is a finite dimensional feature space, and the exact value of $d$ will determined by cross-validation in later experiments. However, compute the embedding is a very challenging task for general graphs: it involves performing an inference in graphical model where we need to integrate out all variables expect $H_i$,~\ie,
\begin{align}
  p(H_i | \cbr{x_i}) = \int_{\Hcal^{V-1}} p(H_i, \cbr{h_j} | \cbr{x_j}) \prod_{j \in \Vcal \setminus i} dh_j.   
\end{align}
Only when the graph structure is a tree, exact computation can be carried out efficiently via message passing~\citep{Pearl88}. Thus in the general case, approximate inference algorithms, \eg, mean field inference and loopy belief propagation (BP), are developed. In many applications, however, these variational inference algorithms exhibit excellent empirical performance~\citep{MurWeiJor99}. Several theoretical studies have also provided insight into the approximations made by loopy BP, partially justifying its application to graphs with cycles~\citep{WaiJor08,YedFreWei01}. 

In the following subsection, we will explain the embedding of mean field and loopy BP. The embedding of other variational inference methods, \eg, double-loop BP, damped BP, tree-reweighted BP, and generalized BP will be explained in Appendix~\ref{appendix:other_embedding}. We show that the iterative update steps in these algorithms, which are essentially minimizing  approximations to the exact free energy, can be simply viewed as function mappings of the embedded marginals using the alternative view in \eq{eq:embed_funtional} and \eq{eq:embed_operator}.

\subsection{Embedding Mean-Field Inference}

The vanilla mean-field inference tries to approximate $p(\cbr{H_i}|\cbr{x_i})$ with a product of \emph{independent} density components $p(\cbr{H_i}|\cbr{x_i}) \approx \prod_{i\in \Vcal} q_i(h_i)$ where each $q_i(h_i) \geq 0$ is a valid density, such that $\int_{\Hcal} q_i(h_i)dh_i = 1$. Furthermore, these density components are found by minimizing the following variational free energy~\citep{WaiJor08}, 
\begin{eqnarray*}
\min_{q_1, \ldots, q_d} \int_{\Hcal^d} \prod_{i\in \Vcal} q_i(h_i) \log\frac{\prod_{i\in \Vcal} q_i(h_i)}{p(\cbr{h_i}|\cbr{x_i})} \prod_{i \in \Vcal} dh_i.
\end{eqnarray*}
One can show that the solution to the above optimization problem needs to satisfy the following fixed point equations for all $i \in \Vcal$

\begin{align}\label{eq:mf_message}
\log q_i(h_i) =& c_i + \log(\Phi(h_i,x_i)) + 
  \sum_{j \in \Ncal(i)} \int_{\Hcal} q_j(h_j) \log(\Psi(h_i,h_j) \Phi(h_j,x_j)) d h_j \nonumber \\
  =&  c_i'+ \log \Phi(h_i, x_i) + \sum_{j\in \Ncal(i)} \int_{\Hcal} q_j(h_j)\log \Psi(h_i, h_j)dh_j \nonumber
\end{align}
where $c_i' = c_i + \sum_{j\in \Ncal(i)}\int q_j(h_j)\log\Phi(h_j, x_j) dh_j$. Here $\Ncal(i)$ are the set of neighbors of variable $H_i$ in the graphical model, and $c_i$ is a constant. The fixed point equations in \eq{eq:mf_message} imply that $q_i(h_i)$ is a functional of a set of neighboring marginals $\cbr{q_j}_{j \in \Ncal(i)}$,~\ie, 
\begin{align}
  q_i(h_i) = f\rbr{h_i, x_i, \cbr{q_j}_{j \in \Ncal(i)}}. 
\end{align}
If for each marginal $q_i$, we have an injective embedding 
$$
\widetilde \mu_i=\int_{\Hcal} \phi(h_i) q_i(h_i) dh_i,
$$
then, using similar reasoning as in \eq{eq:embed_funtional}, we can equivalently express the fixed point equation from an embedding point of view,~\ie,
$q_i(h_i) = \ftil (h_i, x_i, \{\widetilde \mu_j\}_{j \in \Ncal(i)})$, 
and consequently using the operator view from \eq{eq:embed_operator}, we have
\begin{align}
  \widetilde \mu_i = \widetilde \Tcal \circ \rbr{x_i, \cbr{\widetilde \mu_j}_{j \in \Ncal(i)}}. \label{eq:embed_mf} 
\end{align} 
For the embedded mean field~\eq{eq:embed_mf}, the function $\ftil$ and operator $\widetilde \Tcal$ have complicated nonlinear dependencies on the potential functions $\Psi$, $\Phi$, and the feature mapping $\phi$ which is unknown and need to be learned from data. Instead of first learning the $\Psi$ and $\Phi$, and then working out $\widetilde \Tcal$, we will pursue a different route where we directly parameterize $\widetilde \Tcal$ and later learn it with supervision signals.

In terms of the parameterization, we will assume $\widetilde \mu_i \in \RR^d$ where $d$ is a hyperparameter chosen using cross-validation. For $\widetilde \Tcal$, one can use any nonlinear function mappings. For instance, we can parameterize it as a neural network
\begin{align} 
  \widetilde \mu_i &= \sigma\Big(W_1 x_i + W_2 \sum_{j \in \Ncal(i)} \widetilde \mu_j \Big) 
\end{align}
where $\sigma(\cdot):= \max\{0, \cdot\}$ is a rectified linear unit applied elementwisely to its argument, and $\Wb =\{W_1 ,W_2\}$. The number of the rows in $\Wb$ equals to $d$. With such parameterization, the mean field iterative update in the embedding space can be carried out as Algorithm~\ref{alg:mf_msg}. We could also multiply $\widetilde \mu_i$ with $V$ to rescale the range of message embeddings if needed. In fact, with or without $V$, the functions will be the same in terms of the representation power. Specifically, for any $(\Wb, V)$, we can always find another `equivalent' parameters $(\Wb', I)$ where $\Wb' = \{W_1, W_2V\}$.

\begin{figure}[t]
\begin{minipage}{0.45\textwidth}
    \begin{algorithm}[H]
    \caption{\textbf{Embedded Mean Field}}\label{alg:mf_msg}
\begin{algorithmic}[1] 
    \STATE {\bf Input:} parameter $\Wb$ in $\widetilde \Tcal$ 
    \STATE Initialize $\widetilde \mu_i^{(0)} = \zero$, for all $i \in \Vcal$
    \FOR{$t=1$ {\bfseries to} $T$}
    \FOR{$i \in \Vcal$}
    \STATE $l_i = \sum_{j \in \Ncal(i)} \widetilde \mu_j^{(t-1)}$
    \STATE $\widetilde\mu_i^{(t)} =  \sigma(W_1 x_i + W_2 l_i)$
    \ENDFOR 
    \ENDFOR \COMMENT{fixed point equation update}
    \STATE return $\{\widetilde\mu_i^{T}\}_{i \in \Vcal}$
\end{algorithmic}
\end{algorithm}
\end{minipage}
~~~~~~~
\begin{minipage}{0.45\textwidth}
    \begin{algorithm}[H]
    \caption{\textbf{Embedding Loopy BP}}
    \label{alg:lbp_msg}
 \begin{algorithmic}[1] 
    \STATE {\bf Input:} parameter $\Wb$ in $\widetilde \Tcal_1$ and $\widetilde \Tcal_2$
    \STATE Initialize $\widetilde\nu_{ij}^{(0)} = \zero$, for all $(i, j) \in \Ecal$ 
    \FOR{$t=1$ {\bfseries to} $T$}
    \FOR{$(i, j)\in \Ecal$}
    \STATE $\widetilde\nu_{ij}^{t} = \sigma(W_1x_i + W_2 \sum_{k \in \Ncal(i) \setminus j}\, \widetilde\nu_{ki}^{(t-1)})$
    \ENDFOR
    \ENDFOR
    \FOR{$i\in \Vcal$}
    \STATE $\widetilde\mu_i = \sigma(W_3x_i + W_4 \sum_{k \in \Ncal(i) \setminus j}\widetilde\nu_{ki}^{(T)})$
    \ENDFOR
    \STATE return $\{\widetilde\mu_i\}_{i \in \Vcal}$
\end{algorithmic}
\end{algorithm}
\end{minipage}

\end{figure}

\subsection{Embedding Loopy Belief Propagation}

Loopy belief propagation is another variational inference method, which essentially optimizes the Bethe free energy taking \emph{pairwise} interactions into account~\citep{YedFreWei01b},
\begin{eqnarray*}
\hspace{-4mm}&\min_{\{q_{ij}\}_{(i, j)\in \Ecal}}-\sum_{i}(|\Ncal(i)| - 1)\int_\Hcal q_i(h_i)\log\frac{q_i(h_i)}{\Phi(h_i, x_i)}dh_i+\sum_{i, j}\int_{\Hcal^2} q_{ij}(h_i, h_j)\log\frac{q_{ij}(h_i, h_j)}{\Psi(h_i, h_j)\Phi(h_i, x_i)\Phi(h_j, x_j)}dh_i dh_j  \nonumber
\end{eqnarray*}
subject to pairwise marginal consistency constraints: $\int_\Hcal q_{ij}(h_i, h_j)dh_j = q_i(h_i)$, $\int_\Hcal q_{ij}(h_i, h_j)dh_j = q_i(h_i) $, and $\int_\Hcal q_i(h_i) dh_i= 1$. One can obtain the fixed point condition for the above optimization for all
$(i,j) \in \Ecal$,  
\begin{align}\label{eq:lbp_message}
  m_{ij}(h_j) &\propto \int_{\Hcal} \prod_{k\in \Ncal(i)\setminus j}\hspace{-2mm} m_{ki}(h_i)\Phi_i(h_i, x_i) \Psi_{ij}(h_i, h_j)dh_i, \nonumber\\
  q_i(h_i) &\propto \Phi(h_i, x_i)\prod_{j\in \Ncal(i)}m_{ji}(h_i).
\end{align}
where $m_{ij}(h_j)$ is the intermediate result called the message from node $i$ to $j$. Furthermore, $m_{ij}(h_j)$ is a nonnegative function which can be normalized to a density, and hence can also be embedded.

Similar to the reasoning in the mean field case, the~\eq{eq:lbp_message} implies the messages $m_{ij}(h_j)$ and marginals $q_i(h_i)$ are functionals of messages from neighbors, \ie,
\begin{eqnarray*}
m_{ij}(h_j) = f\rbr{h_j, x_i, \{m_{ki}\}_{k\in \Ncal(i)\setminus j}},\\
q_i(h_i) = g\rbr{ h_i, x_i, \{m_{ki}\}_{k\in \Ncal(i)}}.
\end{eqnarray*}
With the assumption that there is an injective embedding for each message $\widetilde \nu_{ij} = \int \phi(h_j) m_{ij}(h_j)dh_j$ and for each marginal 
$\widetilde \mu_{i} = \int \phi(h_i) q_i(h_i) dh_i$, we can apply the reasoning from~\eq{eq:embed_funtional} and ~\eq{eq:embed_operator}, and express the messages and marginals from the embedding view, 
\begin{eqnarray}\label{eq:embed_lbp}
\widetilde\nu_{ij} &=& \widetilde\Tcal_1\circ\rbr{x_i, \cbr{\widetilde \nu_{ki}}_{k\in \Ncal(i)\setminus j}},\\
\widetilde\mu_{i} &=& \widetilde\Tcal_2\circ\rbr{x_i, \cbr{\widetilde \nu_{ki}}_{k\in \Ncal(i)}}.
\end{eqnarray}
We will also use parametrization for loopy BP embedding similar to the mean field case, \ie, neural network with rectified linear unit $\sigma$. Specifically, assume $\widetilde\nu_{ij}\in \RR^d$, $\widetilde\mu_{i}\in \RR^d$
\begin{align}
  \widetilde\nu_{ij} &= \sigma\Big(W_1x_i + W_2\sum_{k\in \Ncal(i)\setminus j}\widetilde\nu_{ki}\Big)\\[-1mm]
  \widetilde\mu_i &=  \sigma\Big(W_3x_i + W_4\sum_{k\in \Ncal(i)}\widetilde\nu_{ki}\Big)
\end{align}
where $\Wb =\{W_1, W_2, W_3, W_4\}$ are matrices with appropriate sizes. Note that one can use other nonlinear function mappings to parameterize $\widetilde\Tcal_1$ and $\widetilde\Tcal_2$ as well. Overall, the loopy BP embedding updates is summarized in Algorithm~\ref{alg:lbp_msg}.

With similar strategy as in mean field case, we will learn the parameters in $\widetilde\Tcal_1$ and $\widetilde\Tcal_2$ later with supervision signals from the discriminative task. 

\subsection{Embedding Other Variational Inference}

In fact, there are many other variational inference methods, with different forms of free energies or different optimization algorithms, resulting different message update forms, \eg, double-loop BP~\citep{Yuille02}, damped BP~\citep{Minka01b}, tree-reweightd BP~\citep{WaiJaaWill03b}, and generalized BP~\citep{YedFreWei01b}. The proposed embedding method is a general technique which can be tailored to these algorithms. The major difference is the dependences in the messages. For the details of embedding of these algorithms, please refer to Appendix~\ref{appendix:other_embedding}. 

\section{Discriminative Training}

Similar to kernel BP~\citep{SonGreGue10,SonGreBicLowGue11} and kernel EP~\citep{JitGreHeeEsletal15}, our current work exploits feature space embedding to reformulate graphical model inference procedures. However, different from the kernel BP and kernel EP, in which the feature spaces are chosen beforehand and the conditional embedding operators are learned locally, our approach will learn both the feature spaces, the transformation $\widetilde \Tcal$, as well as the regressor or classifier for the target values end-to-end using label information. 

Specifically, we are provided with a training dataset $\Dcal = \{\chi_n, y_n\}_{n=1}^N$, where $\chi_n$ is a structured data point and $y_n \hspace{-2mm}\in \hspace{-2mm}\Ycal$, where $\Ycal  = \RR$ for regression or $\Ycal= \{1,\ldots, K\}$ for classification problem, respectively. With the feature embedding procedure introduced in Section~\ref{sec:lvm_embedding}, each data point will be represented as a set of embeddings $\{\tilde \mu_i^n\}_{i \in V_n}\in\Fcal$. Now the goal is to learn a regression or classification function 
$f$ linking $\{\tilde \mu_i^n\}_{i \in V_n}$ to $y_n$. 

More specifically, in the case of regression problem, we will parametrize function $f(\chi_n)$ as $u^\top \sigma(\sum_{i=1}^{V_n} \tilde \mu_i^n)$, where $u \in \RR^d$ is the final mapping from summed (or pooled) embeddings to output. The parameters $u$ and those $\Wb$ involved in the embeddings are learned by minimizing the empirical square loss 
\begin{eqnarray}\label{eq:regression}
\min_{u, \Wb} \sum\nolimits_{n=1}^N \rbr{y_n - u^\top \sigma\rbr{\sum\nolimits_{i=1}^{V_n} \widetilde \mu_i^n}}^2.
\end{eqnarray}
Note that each data point will have its own graphical model and embedded features due to its individual structure, but the parameters $u$ and $\Wb$, are shared across these graphical models. 

In the case of {$K$-class classification} problem, we denote $z$ is the $1$-of-$K$ representation of $y$, \ie, $z\in\{0, 1\}^K$, $z^k = 1$ if $y = k$, and $z^i = 0$, $\forall i\neq k$. By adopt the softmax loss, we obtain the optimization for embedding parameters and discriminative classifier estimation as,
\begin{eqnarray}\label{eq:classification}
\min_{\ub=\{u^k\}_{k=1}^K, \Wb}\sum_n^N\sum_{k=1}^K -z_n^k\log u^k\sigma\rbr{\sum_{i=1}^{V_n} \widetilde \mu_i^n}, 
\end{eqnarray}
where $\ub = \{u^k\}_{k=1}^K$, $u^k\in \RR^d$ are the parameters for mapping embedding to output.

\begin{algorithm}[t]
   \caption{\textbf{Discriminative Embedding}}
   \label{alg:framework}
\begin{algorithmic}
   \STATE {\bf Input:} Dataset $\Dcal = \{\chi_n, y_n\}_{n=1}^N$, loss function $l(f(\chi), y)$.
   \STATE Initialize $\Ub^0 = \{\Wb^0, \ub^0\}$ randomly.
   \FOR{$t=1$ {\bfseries to} $T$}
   \STATE Sample $\{\chi_t, y_t\}$ uniform randomly from $\Dcal$.
   \STATE Construct latent variable model $p(\{H^t_i\}|\chi_n)$ as~\eq{eq:data_model}.
   \STATE Embed $p(\{H^t_i\}|\chi_n)$ as $\cbr{\widetilde\mu_i^n}_{i \in \Vcal_n}$ by Algorithm~\ref{alg:mf_msg} or~\ref{alg:lbp_msg} with ${ \Wb^{t-1}}$.
   \STATE Update $\Ub^t = \Ub^{t-1} +  \lambda_t\nabla_{\Ub^{t-1}} l(f(\widetilde\mu^n; \Ub^{t-1}), y_n)$.
   \ENDFOR
   \STATE return $\Ub^T = \{\Wb^T, \ub^T\}$ 
\end{algorithmic}
\end{algorithm}

The same idea can also be generalized to other discriminative tasks with different loss functions. As we can see from the optimization problems~(\ref{eq:regression}) and (\ref{eq:classification}), the objective functions are directly related to the corresponding discriminative tasks, and so as to $\Wb$ and $\ub$. Conceptually, the procedure starts with representing each datum by a graphical model constructed corresponding to its \emph{individual} structure with \emph{sharing} potential functions, and then, we embed these graphical models with the \emph{same} feature mappings. Finally the embedded marginals are aggregated with a prediction function for a discriminative task. The shared potential functions, feature mappings and final prediction functions are all learned together for the ultimate task with supervision signals.

We optimize the objective~\eq{eq:regression} or~\eq{eq:classification} with stochastic gradient descent for scalability consideration. However, other optimization algorithms are also applicable, and our method does not depend on this particular choice. The gradients of the parameters $\Wb$ are calculated recursively similar to recurrent neural network for sequence models. In our case, the recursive structure will correspond the message passing structure. The overall framework is illustrated in Algorithm~\ref{alg:framework}. For details of the gradient calculation, please refer to Appendix~\ref{appendix:derivative}.
	
\section{Experiments}

Below we first compare our method with algorithms using prefixed kernel on string and graph benchmark datasets. Then we focus on Harvard Clean Energy Project dataset which contains 2.3 million samples. We demonstrate that while getting comparable performance on medium sized datasets, we are able to handle millions of samples, and getting much better when more training data are given. The two variants of {\tt structure2vec} are denoted as DE-MF and DE-LBP, which stands for discriminative embedding using mean field or loopy belief propagation, respectively. 

Our algorithms are implemented with C++ and CUDA, and experiments are carried out on clusters equipped with NVIDIA Tesla K20. The code is available on \href{https://github.com/Hanjun-Dai/graphnn}{https://github.com/Hanjun-Dai/graphnn}.

\subsection{Benchmark structure datasets}

We compare our algorithm on string benchmark datasets with the kernel method with existing sequence kernels, \ie, the spectrum string kernel~\citep{LesEskNob02}, mismatch string kernel~\citep{LesEskWesNob02} and fisher kernel with HMM generative models~\citep{JaaHau99b}. On graph benchmark datasets, we compare with subtree kernel~\citep{RamGae03} (R\&G, for short), random walk kernel\citep{GaeFlaWro03, VisSchKonBor10}, shortest path kernel~\citep{BorKri05}, graphlet kernel\citep{SheVisPetMehetal09} and the family of Weisfeiler-Lehman kernels (WL kernel)~\citep{SheSchVanMehetal11}. After getting the kernel matrix, we train SVM classifier or regressor on top. 

We tune all the methods via cross validation, and report the average performance. Specifically, for structured kernel methods, we tune the degree in $\{1, 2, \ldots, 10\}$ (for mismatch kernel, we also tune the maximum mismatch length in $\{1, 2, 3\}$) and train SVM classifier~\citep{ChaLin01b} on top, where the trade-off parameter $C$ is also chosen in $\{0.01, 0.1, 1, 10\}$ by cross validation. For fisher kernel that using HMM as generative model, we also tune the number of hidden states assigned to HMM in $\{2, \ldots, 20\}$.

For our methods, we simply use one-hot vector (the vector representation of discrete node attribute) as the embedding for observed nodes, and use a two-layer neural network for the embedding (prediction) of target value. The hidden layer size $b \in \{16, 32, 64\}$ of neural network, the embedding dimension $d \in \{16, 32, 64\}$ of hidden variables and the number of iterations $t \in \{1, 2, 3, 4\}$ are tuned via cross validation. We keep the number of parameters small, and use early stopping~\citep{CarLawGil01} to avoid overfitting in these small datasets. 
  
\subsubsection{String Dataset}

Here we do experiments on two string binary classification benchmark datasets. The first one (denoted as SCOP) contains 7329 sequences obtained from SCOP (Structural Classification of Proteins) 1.59 database~\citep{AndHowBreHubetal04}. Methods are evaluated on the ability to detect members of a target SCOP family (positive test set) belonging to the same SCOP superfamily as the positive training sequences, and no members of the target family are available during training. We use the same 54 target families and the same training/test splits as in remote homology detection~\citep{KuaIeWanWanetal05}. The second one is FC and RES dataset (denoted as FC\_RES) provided by CRISPR/Cas9 system, on which the task it to tell whether the guide RNA will direct Cas9 to target DNA. There are 5310 guides included in the dataset. Details of this dataset can be found in \citet{DoeHarGraTotetal14, FusSmiDoeLis15}. 
We use two variants for spectrum string kernel: 1) kmer-single, where the constructed kernel matrix $K_k^{(s)}$ only consider patterns of length $k$; 2) kmer-concat, where kernel matrix $K^{(c)} = \sum_{i=1}^k K_k^{(s)}$. We also find the normalized kernel matrix $K_k^{Norm}(x, y) = \frac{K_k(x, y)}{\sqrt{K_k(x, x)K_k(y, y)}}$ helps.

\begin{table}[h]
\begin{center}
\begin{tabular}{l|c|c}
 & FC\_RES & SCOP \\
\hline
kmer-single & 0.7606$\pm$0.0187  & 0.7097$\pm$0.0504 \\
\hline
kmer-concat & 0.7576$\pm$0.0235  & 0.8467$\pm$0.0489 \\
\hline
mismatch & 0.7690$\pm$0.0197 & 0.8637$\pm$0.1192\\
\hline
fisher & 0.7332$\pm$0.0314 & 0.8662$\pm$0.0879 \\
\hline
DE-MF & \textbf{0.7713$\pm$0.0208} & 0.9068$\pm$0.0685 \\
\hline
DE-LBP & 0.7701$\pm$0.0225 & \textbf{0.9167$\pm$0.0639} \\
\hline
\end{tabular}
\end{center}
\vspace{-3mm}
\caption{Mean AUC on string classification datasets}
\label{tab:str_auc}
\vspace{-3mm}
\end{table}

Table~\ref{tab:str_auc} reports the mean AUC of different algorithms. 
We found two variants of {\tt structure2vec} are consistently better than the string kernels. Also, the improvement in SCOP is more significant than in FC\_RES. This is because SCOP is a protein dataset and its alphabet size $|\Sigma|$ is much larger than that of FC\_RES, an RNA dataset. Furthermore, the dimension of the explicit features for a k-mer kernel is $O(|\Sigma|^k)$, which can make the off-diagonal entries of kernel matrix very small (or even zero) with large alphabet size and $k$. That's also the reason why kmer-concat performs better than kmer-single. {\tt structure2vec} learns a discriminative feature space, rather than prefix it beforehand, and hence does not have this problem. 

\subsubsection{Graph Dataset}

 \begin{figure*}[t] 
 \small 
 \centering 
 \includegraphics[scale=0.4]{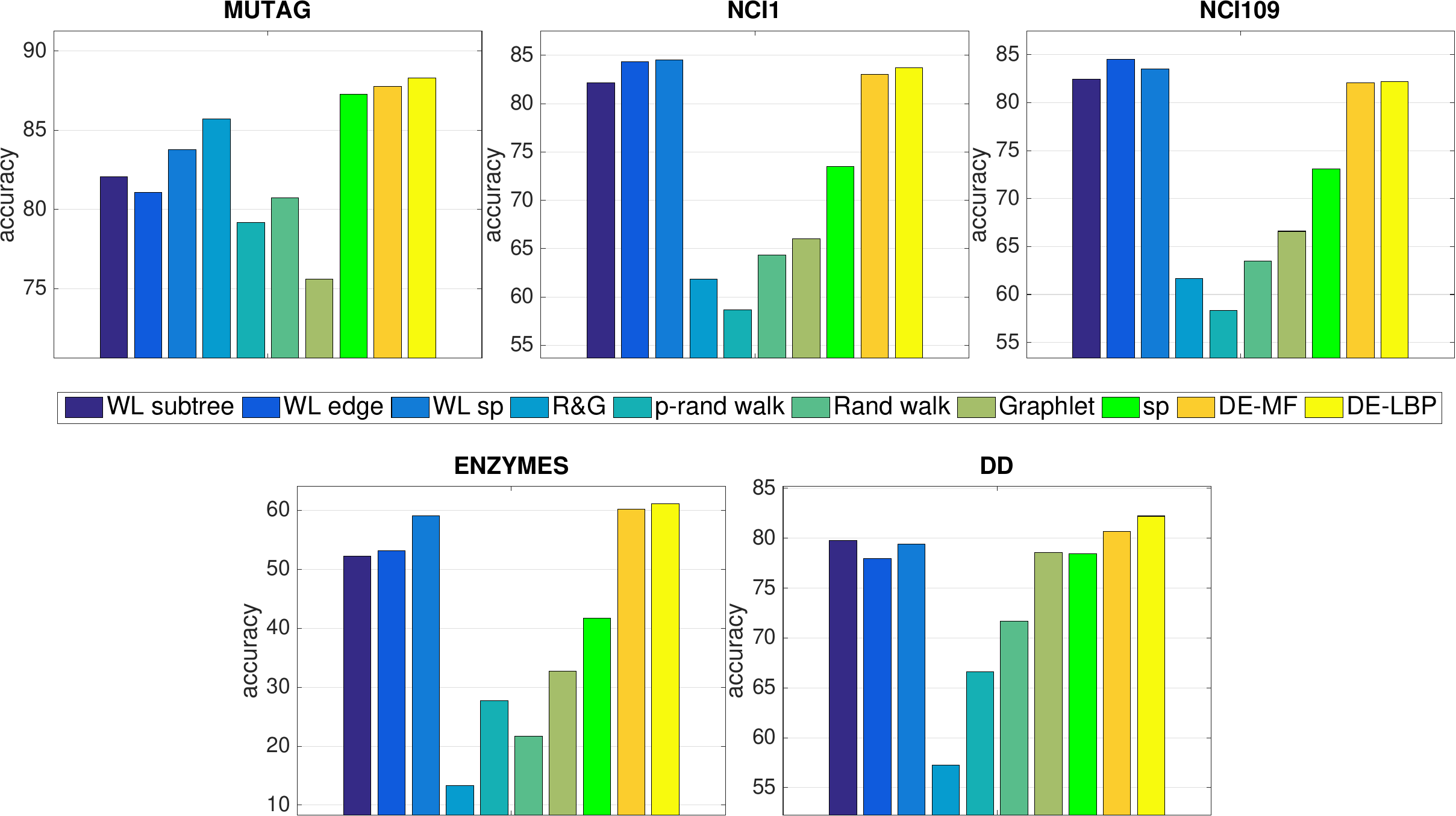}
 \caption{10-fold cross-validation accuracies on graph classification benchmark datasets. The `sp' in the figure stands for shortest-path. For corresponding numerical results, see Table~\ref{tab:graph_kernel_results} }
 \label{fig:gk_acc}
 \end{figure*}

We test the algorithms on five benchmark datasets for graph kernel:  MUTAG, NCI1, NCI109, ENZYMES and D\&D. MUTAG~\citep{DebLopDebShuetal91}. NCI1 and NCI109~\citep{WalWatKar08} are chemical compounds dataset, while ENZYMES~\citep{BorKri05} and D\&D~\citep{DobDoi03} are of proteins. The task is to do multi-class or binary classification. We show the detailed statistics of these datasets in Table~\ref{tab:graph_stats}. 

\begin{table}
\begin{center}
  \begin{tabular}{|l|c|c|c|c|}
    \hline
    & size & avg $|V|$ & avg $|E|$ & \#labels \\
    \hline
    MUTAG & 188 & 17.93 & 19.79 & 7\\
    \hline
    NCI1 & 4110 & 29.87 & 32.3 & 37 \\
    \hline
    NCI109 & 4127 & 29.68 & 32.13 & 38 \\
    \hline
    ENZYMES & 600 & 32.63 & 62.14 & 3 \\
    \hline
    D\&D & 1178 & 284.32 & 715.66 & 82 \\
    \hline
  \end{tabular}
\end{center}
  \caption{Statistics~\citep{SugBor15} of graph benchmark datasets. $|V|$ is the \# nodes while $|E|$ is the \# edges in a graph. \#labels equals to the number of different types of nodes. }
  \label{tab:graph_stats}
\end{table}

The results of baseline algorithms are taken from~\citet{SheSchVanMehetal11} since we use exactly the same setting here. From the accuracy comparison shown in Figure~\ref{fig:gk_acc}, we can see the proposed embedding methods are comparable to the alternative graph kernels, on different graphs with different number of labels, nodes and edges. Also, in dataset D\&D which consists of 82 different types of labels, our algorithm performs much better. As reported in~\citet{SheSchVanMehetal11}, the time required for constructing dictionary for the graph kernel can take up to more than a year of CPU time in this dataset, while our algorithm can learn the discriminative embedding efficiently from structured data directly without the construction of the handcraft dictionary. 

\subsection{Harvard Clean Energy Project(CEP) dataset}

\begin{figure}[t] 
 \small 
 \centering 
 \subfigure[b][PCE distribution]
 {
  \includegraphics[width=0.38\linewidth]{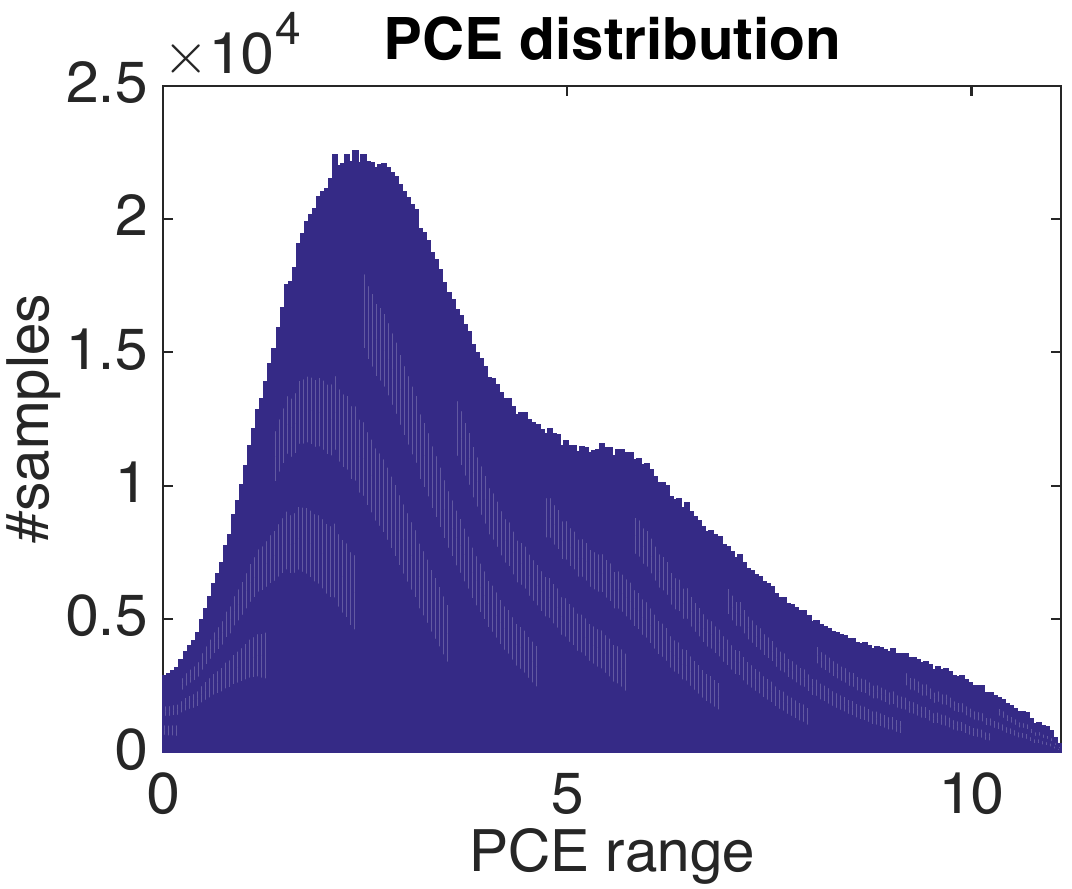}
 }
 \hspace{1cm}
 \subfigure[b][Sample molecules]
 {
   \includegraphics[width=0.38\linewidth]{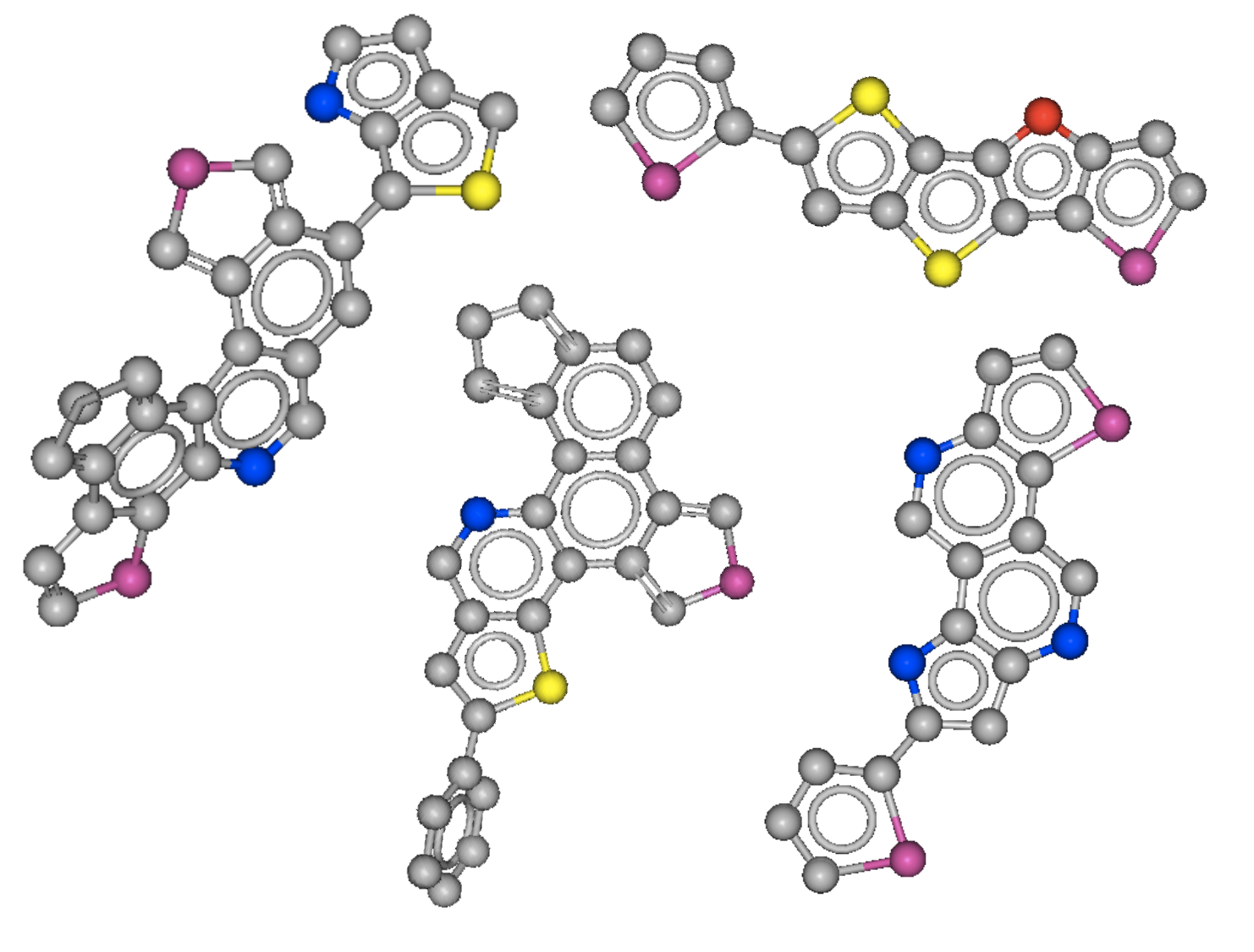}
 }
 \caption{PCE value distribution and sample molecules from CEP dataset. Hydrogens are not displayed. }
 \label{fig:cep}
 \end{figure}

The Harvard Clean Energy Project~\citep{HacOliAtaAmaetal11} is a theory-driven search for the next generation of organic solar cell materials. One of the most important properties of molecule for this task is the overall efficiency of the energy conversion process in a solar cell, which is determined by the power conversion efficiency (PCE). The Clean Energy Project (CEP) performed expensive simulations for the 2.3 million candidate molecules on IBM’s World Community Grid, in order to get this property value. So using machine learning approach to accurately predict the PCE values is a promising direction for the high throughput screening and discovering new materials. 

In this experiment, we randomly select 90\% of the data for training, and the rest 10\% for testing. This setting is similar to~\citet{PyzLiAsp15}, except that we use the entire 2.3m dataset here. Since the data is distributed unevenly (see Figure~\ref{fig:cep}), we resampled the training data (but not the test data) to make the algorithm put more emphasis on molecules with higher PCE values, in order to make accurate prediction for promising candidate molecules.  
Since the traditional kernel methods are not scalable, we make the explicit feature maps for WL subtree kernel by collecting all the molecules and creating dictionary for the feature space. The other graph kernels, like edge kernel and shortest path kernel, are having too large feature dictionary to work with. We use RDKit~\citep{Landrum12} to extract features for atoms (nodes) and bonds (edges).
 
 The mean absolute error (MAE) and root mean square error (RMSE) are reported in Table~\ref{tab:cep_test}. We found utilizing graph information can accurately predict PCE values. Also, our proposed two methods are working equally well. Although WL tree kernel with degree 6 is also working well, it requires $10,000$ times more parameters than {\tt structure2vec} and runs 2 times slower. The preprocessing needed for WL tree kernel also makes it difficult to use in large datasets. 
  
\begin{figure}[t] 
 \small 
 \centering 
 \subfigure[b][Test error vs iterations]
 {
 \includegraphics[width=0.38\linewidth]{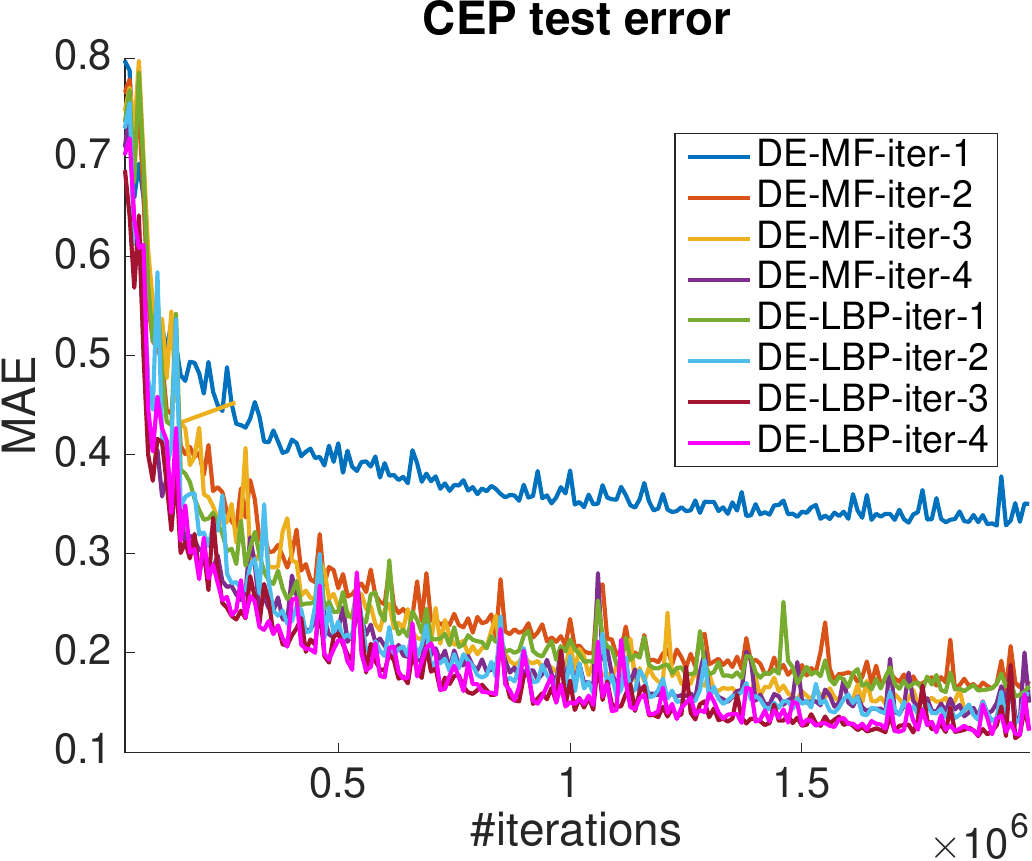}
 }
 \hspace{1cm}
 \subfigure[b][Prediction quality]
 {
 \includegraphics[width=0.38\linewidth]{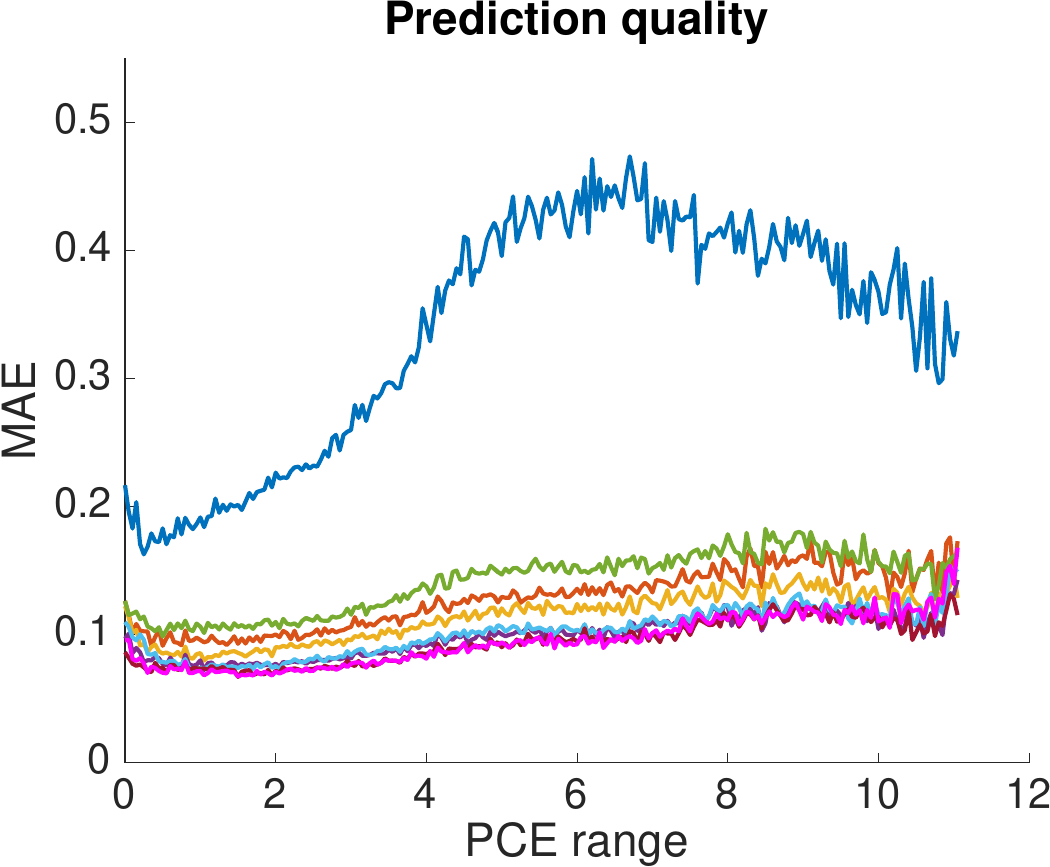}
 }
 \caption{Details of training and prediction results for DE-MF and DE-LBP with different number of fixed point iterations. }
 \label{fig:cep_exp_detail}
 \end{figure}
 
\begin{table}[h]
\begin{center}
\begin{tabular}{l|c|c|c}
 & test MAE & test RMSE & \# params \\
\hline
Mean Predictor & 1.9864 & 2.4062 & 1 \\
\hline
WL lv-3 & 0.1431 & 0.2040 & 1.6m\\
\hline
WL lv-6 & 0.0962 & 0.1367 & 1378m\\
\hline
DE-MF  & 0.0914 & 0.1250 & 0.1m \\
\hline
DE-LBP & \textbf{0.0850} & \textbf{0.1174} & 0.1m\\
\hline
\end{tabular}
\end{center}
\caption{Test prediction performance on CEP dataset. WL lv-$k$ stands for Weisfeiler-lehman with degree $k$. }
\label{tab:cep_test}
\end{table}

To understand the effect of the inference embedding in the proposed algorithm framework, we further compare our methods with different number of fixed point iterations in Figure~\ref{fig:cep_exp_detail}. It can see that, higher number of fixed point iterations will lead to faster convergence, though the number of parameters of the model in different settings are the same. The mean field embedding will get much worse result if only one iteration is executed. Compare to the loopy BP case with same setting, the latter one will always have one more round message passing since we need to aggregate the messages from edge to node in the last step. And also, from the quality of prediction we find that, though making slightly higher prediction error for molecules with high PCE values due to insufficient data, the variants of our algorithm are not overfitting the `easy' (i.e., the most popular) range of PCE value.    

\section{Conclusion}

We propose, {\tt structure2vec}, an effective and scalable approach for structured data
representation based on the idea of embedding latent variable models into feature spaces, and learning such feature spaces using discriminative information. Interestingly, {\tt structure2vec} extracts features by performing a sequence of function mappings in a way similar to graphical model inference procedures, such
as mean field and belief propagation. In applications involving millions of data points, we showed that {\tt structure2vec} runs 2 times faster, produces models $10,000$ times smaller, while at the same time achieving the state-of-the-art predictive performance. {\tt structure2vec} provides a nice example for the general strategy of combining the strength of graphical models, Hilbert space embedding of distribution and deep learning approach, which we believe will become common in many other learning tasks.

{\bf Acknowledgements.} This project was supported in part by NSF/NIH BIGDATA 1R01GM108341, ONR N00014-15-1-2340, NSF IIS-1218749, and NSF CAREER IIS-1350983.

\clearpage
\newpage
\onecolumn

\begin{center}
{\huge Appendix}
\end{center}

\begin{appendix}

\section{More Related Work}
\label{sec:more_related_work} 

 \subsection{Comparison with Neural Networks on Graphs}

Neural network is also a powerful tool on graph structured data. \citet{ScaGorTsoHagetal09} proposed a neural network which generates features by solving a heuristic nonlinear system iteratively, and is learned using Almeida-Pineda algorithm. To guarantee the existence of the solution to the nonlinear system, there are extra requirements for the features generating function. From this perspective, the model in~\citep{LiTarBroZem15} can be considered as an extension of~\citep{ScaGorTsoHagetal09} where the gated recurrent unit is used for feature generation. Rather than these heuristic models, our model is based on the principled graphical model embedding framework, which results in flexible embedding functions for generating features. Meanwhile, the model can be learned efficiently by traditional stochastic gradient descent. 

There are several work transferring locality concept of convolutional neural networks~(CNN) from Euclidean domain to graph case, using hierarchical clustering, graph Laplacian~\citep{BruZarSzlLeC13}, or graph Fourier transform \citep{HenBruLeC15}. These models are still restricted to problems with the same graph structure, which is not suitable for learning with molecules. \cite{MouLiZhaWanetal16} proposed a convolution operation on trees, while the locality are defined based on parent-child relations. \cite{DuvMacIpaBometal15} used CNN to learn the circulant fingerprints for graphs from end to end. The dictionary of fingerprints are maintained using softmax of subtree feature representations, in order to obtain a differentiable model. If we unroll the steps in Algorithm~\ref{alg:framework}, it can also be viewed as an end to end learning system. However, the structures of the proposed model are deeply rooted in graphical model embedding, from mean field and loopy BP, respectively. Also, since the parameters will be shared 
across different unrolling steps, we would have more compact model. As will be shown in the experiment section, our model is easy to train, while yielding good generalization ability. 

 \subsection{Comparison with Learning Message Estimator}

By recognizing inference as computational expressions, inference machines~\citep{RosMunHebBag11} incorporate learning into the messages passing inference for CRFs. More recently, \citet{HerRouWen14, ZheJayRomVinetal15, LinSheReiHen15} designed specific recurrent neural networks and convolutional neural networks for imitating the messages in CRFs. Although these methods share the similarity, \ie, bypassing learning potential function, to the proposed framework, there are significant differences comparing to the proposed framework. 

The most important difference lies in the learning setting. In these existing messages learning work~\citep{HerRouWen14, ZheJayRomVinetal15, LinSheReiHen15, CheSchYuiUrt14}, the learning task is still estimating the messages represented graphical models with designed function forms, \eg, RNN or CNN, by maximizing loglikelihood. While in our work, we represented each structured data as a distribution, and the learning task is regression or classification over these {distributions}. Therefore, we treat the embedded models as samples, and learn the nonlinear mapping for embedding, and regressor or classifier, $f:\Pcal\rightarrow\Ycal$, over these distributions jointly, with task-dependent user-specified loss functions. 

Another difference is the way in constructing the messages forms, and thus, the neural networks architecture. In the existing work, the neural networks forms are constructed \emph{strictly} follows the message updates forms~\eq{eq:mf_message} or~\eq{eq:lbp_message}. Due to such restriction, these works only focus on discrete variables with finite values, and is difficult to extend to continuous variables because of the integration. However, by exploiting the embedding point of view, we are able to build the messages with more \emph{flexible} forms without losing the dependencies. Meanwhile, the difficulty in calculating integration for continuous variables is no longer a problem with the reasoning~\eq{eq:embed_funtional} and \eq{eq:embed_operator}. 

\section{Derivation of the Fixed-Point Condition for Mean-Field Inference}\label{appendix:mf_derivation}

In this section, we derive the fixed-point equation for mean-field inference in Section~\ref{sec:lvm_embedding}. As we introduced, the mean-field inference is indeed minimizing the variational free energy,
\begin{eqnarray*}
\min_{q_1, \ldots, q_d} L(\{q_i\}_{i=1}^d) := \int_{\Hcal^d} \prod_{i\in \Vcal} q_i(h_i) \log\frac{\prod_{i\in \Vcal} q_i(h_i)}{p(\cbr{h_i}|\cbr{x_i})} \prod_{i \in \Vcal} dh_i.
\end{eqnarray*}
Plug the MRF~\eq{eq:data_model} into objective, we have
\begin{eqnarray}
L(\{q_i\}_{i=1}^d) \hspace{-2mm} &=& \hspace{-3mm} - \hspace{-2mm}\int_{\Hcal^d}\hspace{-2mm} \prod_{i\in \Vcal} q_i(h_i)\rbr{\log \Phi(h_i, x_i) \hspace{-1mm}+ \hspace{-2mm}\sum_{j\in \Ncal(i)}\log \rbr{\Psi(h_i, h_j)\Phi(h_j, x_j)} \hspace{-1mm}+\hspace{-2mm} \sum_{k\notin \Ncal(i)}\hspace{-2mm}\log\rbr{\prod_{(k, l)\in \Ecal}\Psi(h_k, h_l)\Phi(h_k, x_k)}}\prod_{i\in \Vcal}dh_i \nonumber \\ 
&+&\sum_{i\in \Vcal}\int_{\Hcal} q_i(h_i)\log q_i(h_i) dh_i \nonumber \\
&=& - \int q_i(h_i)\log \Phi(h_i, x_i)dh_i - \sum_{j\in \Ncal(i)}\int q_i(h_i) \rbr{\int q_j(h_j)\log \rbr{\Psi(h_i, h_j)\Phi(h_j, x_j)} dh_j}dh_i + c_i \\
&+& \sum_{i\in \Vcal}\int_{\Hcal} q_i(h_i)\log q_i(h_i) dh_i, \nonumber
\end{eqnarray}
where $c_i = - \int \prod_{k\notin\Ncal(i)}q_k(h_k)\rbr{\sum_{k\notin \Ncal(i)}\log\rbr{\prod_{(k, l)\in \Ecal}\Psi(h_k, h_l)\Phi(h_k, x_k)}}\prod_{k\notin \Ncal(i)}dh_k$. Take functional derivatives of $L(\{q_i\}_{i=1}^d)$ w.r.t. $q_i(h_i)$, and set them to zeros, we obtain the fixed-point condition in Section~\ref{sec:lvm_embedding},
\begin{eqnarray}
\log q_i(h_i)  = c_i + \log \Phi(h_i, x_i) + \sum_{j\in \Ncal(i)} \int q_j(h_j)\log \rbr{\Psi(h_i, h_j)\Phi(h_j, x_j)} dh_j.
\end{eqnarray}

This fixed-point condition could be further reduced due to the independence between $h_i$ and $x_j$ given $h_j$, \ie, 
\begin{eqnarray}
\log q_i(h_i) &=& c_i + \log \Phi(h_i, x_i) + \sum_{j\in \Ncal(i)} \int q_j(h_j)\log \Psi(h_i, h_j)dh_j + \sum_{j\in \Ncal(i)}\int q_j(h_j)\log\Phi(h_j, x_j) dh_j\\
& = & c_i'+ \log \Phi(h_i, x_i) + \sum_{j\in \Ncal(i)} \int q_j(h_j)\log \Psi(h_i, h_j)dh_j,
\end{eqnarray}
where $c_i' = c_i + \sum_{j\in \Ncal(i)}\int q_j(h_j)\log\Phi(h_j, x_j) dh_j$. 

\section{Derivation of the Fixed-Point Condition for Loopy BP}\label{appendix:lbp_derivation}

The derivation of the fixed-point condition for loopy BP can be found in~\cite{YedFreWei01b}. However, to keep the paper self-contained, we provide the details here. The objective of loopy BP is 
\begin{eqnarray*}
\min_{\{q_{ij}\}_{(i, j)\in \Ecal}}&&-\sum_{i}(|\Ncal(i)| - 1)\int_\Hcal q_i(h_i)\log\frac{q_i(h_i)}{\Phi(h_i, x_i)}dh_i+\sum_{i, j}\int_{\Hcal^2} q_{ij}(h_i, h_j)\log\frac{q_{ij}(h_i, h_j)}{\Psi(h_i, h_j)\Phi(h_i, x_i)\Phi(h_j, x_j)}dh_i dh_j\\
\text{s.t.} \quad && \int_\Hcal q_{ij}(h_i, h_j)dh_j = q_i(h_i), \quad \int_\Hcal q_{ij}(h_i, h_j)dh_j = q_i(h_i),\quad \int_\Hcal q_i(h_i) dh_i= 1.
\end{eqnarray*}

Denote $\lambda_{ij}(h_j)$ is the multiplier to marginalization constraints $\int_\Hcal q_{ij}(h_i, h_j)dh_i - q_j(h_j) = 0$, the Lagrangian is formed as 
\begin{eqnarray*}
L(\{q_{ij}\}, \{q_i\}, \{\lambda_{ij}\}, \{\lambda_{ji}\}) &&= -\sum_{i}(|\Ncal(i)| - 1)\int_\Hcal q_i(h_i)\log\frac{q_i(h_i)}{\Phi(h_i, x_i)}dh_i\\
&&+\sum_{i, j}\int_{\Hcal^2} q_{ij}(h_i, h_j)\log\frac{q_{ij}(h_i, h_j)}{\Psi(h_i, h_j)\Phi(h_i, x_i)\Phi(h_j, x_j)}dh_i dh_j \\
&&- \sum_{i,j}\int_\Hcal\lambda_{ij}(h_j)\Big( \int_\Hcal q_{ij}(h_i, h_j)dh_i - q_j(h_j)\Big)dh_j\\
&&- \sum_{i,j}\int_\Hcal\lambda_{ji}(h_i)\Big( \int_\Hcal q_{ij}(h_i, h_j)dh_j - q_i(h_i)\Big)dh_i
\end{eqnarray*}
with normalization constraints $\int_\Hcal q_i(h_i) dh_i= 1$. Take functional derivatives of $L(\{q_{ij}\}, \{q_i\}, \{\lambda_{ij}\}, \{\lambda_{ji}\})$ with respect to $q_{ij}(h_i, h_j)$ and $q_i(h_i)$, and set them to zero, we have
\begin{eqnarray*}
q_{ij}(h_i, h_j) &\propto& \Psi(h_i, h_j)\Phi(h_i, x_i)\Phi(h_j, x_j)\exp(\lambda_{ij}(h_j) + \lambda_{ji}(h_i)), \\
q_i(h_i) &\propto& \Phi(h_i, x_i)\exp\rbr{\frac{\sum_{k\in \Ncal(i)}\lambda_{ki}(h_i)}{|\Ncal(i)| - 1}}.
\end{eqnarray*}
We set $m_{ij}(h_j) = \frac{q_j(h_j)}{\Phi(h_i, x_i)\exp(\lambda_{ij}(h_j))}$, therefore, 
$$
\prod_{k\in \Ncal(i)} m_{ki}(h_i) \propto \exp\rbr{\frac{\sum_{k\in \Ncal(i)}\lambda_{ki}(h_i)}{|\Ncal(i)| - 1}}.
$$ 
Plug it into $q_{ij}(h_i, h_j)$ and $q_i(h_i)$, we recover the loopy BP update for marginal belief and  
$$
\exp(\lambda_{ji}(h_i)) = \frac{q_i(h_i)}{\Phi(h_i, x_i)m_{ji}(h_i)} \propto \prod_{k\in N(i)\setminus j} m_{ki}(h_i).
$$
The update rule for message $m_{ij}(h_j)$ can be recovered using the marginal consistency constraints, 
\begin{eqnarray*}
m_{ij}(h_j) &=& \frac{q_j(h_j)}{\Phi(h_i, x_i)\exp(\lambda_{ij}(h_j))} = \frac{\int_\Hcal q_{ij}(h_i, h_j)dh_i}{\Phi(h_i, x_i)\exp(\lambda_{ij}(h_j))} \\
&=& \Phi(h_j, x_j)\exp(\lambda_{ij}(h_j))\frac{\int_\Hcal \Psi(h_i, h_j)\Phi(h_i, x_i)\exp(\lambda_{ji}(h_i))dh_i}{\Phi(h_i, x_i)\exp(\lambda_{ij}(h_j))} \\
&\propto& \int_\Hcal \Psi(h_i, h_j)\Phi(h_i, x_i)\prod_{k\in N(i)\setminus j} m_{ki}(h_i)dh_i.
\end{eqnarray*}

Moreover, we also obtain the other important relationship between $m_{ij}(h_j)$ and $\lambda_{ji}(h_i)$ by marginal consistency constraint and the definition of $m_{ij}(h_j)$, 
\begin{eqnarray*}
m_{ij}(h_j) \propto \int \Psi(h_i, h_j)\Phi(h_j, x_j)\exp(\lambda_{ji}(h_i))dh_i.
\end{eqnarray*}

\section{Embedding Other Variational Inference}\label{appendix:other_embedding}

The proposed embedding is a general algorithm and can be tailored to other variational inference methods with message passing paradigm. In this section, we discuss the embedding for several alternatives, which optimize the primal and dual Bethe free energy, its convexified version and Kikuchi free energy, respectively. We will parametrize the messages with the same function class, \ie, neural networks with ReLU. More generally, we can also treat the weights as parameters and learn them together.



\subsection{Double-Loop BP}

Noticed the Bethe free energy can be decomposed into the summation of a convex function and a concave function, \citet{Yuille02} utilizes CCCP to minimize the Bethe free energy, resulting the double-loop algorithm. Take the gradient of Lagrangian of the objective function, and set to zero, the primal variable can be represented in dual form,
\begin{eqnarray*}
q_{ij}(h_i, h_j)\propto \Psi(h_i, h_j)\Phi(h_i, x_i)\Phi(h_j, x_j)\exp(\lambda_{ij}(h_j) + \lambda_{ji}(h_i)),\\
q_i(h_i) \propto \Phi(h_i, x_i)\exp\Big(|\Ncal(i)|\gamma_s(h_i) - \sum_{k\in \Ncal(i)}\lambda_{ki}(h_i)\Big),
\end{eqnarray*}
The algorithm updates $\gamma$ and $\lambda$ alternatively,
\begin{eqnarray*}
2\lambda_{ij}^{new}(h_j) &=& |\Ncal(j)|\gamma_i(h_i) - \sum_{{k\in \Ncal(j)\setminus i}} \lambda_{kj}(h_j) - \log \int_\Hcal \Psi(h_i, h_j)\Phi(h_i, x_i) \lambda_{ji}(h_i)d h_i,\\
\gamma_i^{new}(h_i) &=& |\Ncal(i)|\gamma_i(h_i) - \sum_{{k\in \Ncal(i)}}\lambda_{ki}(h_i).
\end{eqnarray*}
Consider the $\lambda_{ij}$ as messages, with the injective embedding assumptions for corresponding messages, follow the same notation, we can express the messages in embedding view
\begin{eqnarray*}
\widetilde\nu_{ij} &=& \widetilde\Tcal_1\circ\rbr{x_i, \cbr{\widetilde \nu_{ki}}_{k\in \Ncal(i)\setminus j}, \widetilde \nu_{ji}, \widetilde\mu_i},\quad \text{or} \quad \widetilde\nu_{ij} = \widetilde\Tcal_1\circ\rbr{x_i, \cbr{\widetilde \nu_{ki}}_{k\in \Ncal(i)}, \widetilde\mu_i}, \\
\widetilde\mu_{i} &=& \widetilde\Tcal_2\circ\rbr{x_i, \cbr{\widetilde \nu_{ki}}_{k\in \Ncal(i)}, \widetilde\mu_i}.
\end{eqnarray*}
Therefore, we have the parametrization form as 
\begin{eqnarray*}
\widetilde\nu_{ij} &=& \sigma\rbr{W_1x_i + W_2\sum_{k\in \Ncal(i)\setminus j} \widetilde \nu_{ki} + W_3 \widetilde \nu_{ji} + W_4 \widetilde\mu_i},\quad \text{or} \quad \widetilde\nu_{ij} = \sigma\rbr{W_1x_i +  W_2\sum_{k\in \Ncal(i)}\widetilde \nu_{ki} + W_3 \widetilde\mu_i}, \\
\widetilde\mu_{i} &=& \sigma\rbr{W_5x_i + W_6\sum_{k\in \Ncal(i)}\widetilde \nu_{ki} + W_7 \widetilde\mu_i}.
\end{eqnarray*}
where $\widetilde\mu_{i}\in \RR^d$, $\widetilde \nu_{ij}\in \RR^d$, and $\Wb = \{W_i\}$ are matrices with appropriate size.

\subsection{Damped BP}

Instead of the primal form of Bethe free energy, \citet{Minka01b} investigates the duality of the optimization,
\begin{eqnarray*}
\min_{\gamma}\max_{\lambda}&&\sum_{i}\Big(|\Ncal(i)| - 1\Big)\log\int_\Hcal\Phi(h_i, x_i)\exp(\gamma_i(h_i))dh_i \\
&-& \sum_{(i, j)\in \Ecal}\log\int_{\Hcal^2}\Psi(h_i, h_j)\Phi(h_i, x_i)\Phi(h_j, x_j)\exp(\lambda_{ij}(h_j) + \lambda_{ji}(h_i))dh_jdh_i,
\end{eqnarray*}
subject to $\Big(|\Ncal(i)| - 1\Big)\gamma_i(h_i) = \sum_{k\in \Ncal(i)}\lambda_{ki}(h_i)$. Define message as 
$$
m_{ij}(h_j) \propto \int_\Hcal \Psi(h_i, h_j)\Phi(h_j, x_j)\exp(\lambda_{ji}(h_i))dh_i,
$$
the messages updates are
\begin{eqnarray*}
m_{ij}(h_i) &\propto& \int_{\Hcal} \Phi_i(h_i, x_i)\Psi_{ij}(h_i, h_j)\exp\Big(\frac{|\Ncal(i)|-1}{|\Ncal(i)|}\gamma_i(h_i)\Big)\frac{\prod_{k\in \Ncal(i)}m_{ki}^\frac{1}{|\Ncal(i)|}(h_i)} {m_{ji}(h_i)}dh_i,\\
\gamma_i^{new}(h_i) &\propto& \frac{|\Ncal(i)|-1}{|\Ncal(i)| }\gamma_i(h_i) + \sum_{k\in \Ncal(i)} {\frac{1}{|\Ncal(i)|}}\log m_{ki}(h_i).
\end{eqnarray*}
and the 
$$
q(x_i) \propto \Phi(h_i, x_i)\exp\rbr{\frac{|\Ncal(i)|-1}{|\Ncal(i)| }\gamma_i(h_i)}\prod_{k\in \Ncal(i)}m_{ki}^\frac{1}{|\Ncal(i)|}
$$
which results the embedding follows the same injective assumption and notations, 
\begin{eqnarray*}
\widetilde \nu_{ij} &=& \widetilde\Tcal_1\circ\rbr{x_i, \cbr{\widetilde \nu_{ki}}_{k\in \Ncal(i)}, \widetilde \nu_{ji}, \widetilde\mu_i},\\
\widetilde \mu_i &=& \widetilde\Tcal_2\circ\rbr{x_i, 	\widetilde \mu_i, \cbr{\widetilde \nu_{ki}}_{k\in \Ncal(i)}}.
\end{eqnarray*}
and the parametrization, 
\begin{eqnarray*}
\widetilde \nu_{ij} &=& \sigma\rbr{W_1x_i + W_2\sum_{k\in \Ncal(i)}{\widetilde \nu_{ki}} + W_3\widetilde \nu_{ji} + W_4 \widetilde\mu_i},\\
\widetilde \mu_i &=& \sigma\rbr{W_5x_i + W_6\widetilde \mu_i + W_7\sum_{k\in \Ncal(i)}{\widetilde \nu_{ki}}}.
\end{eqnarray*}

It is interesting that after parametrization, the embeddings of double-loop BP and damped BP are essentially the same, which reveal the connection between double-loop BP and damped BP.

\subsection{Tree-reweighted BP}

Different from loopy BP and its variants which optimizing the Bethe free energy, the tree-reweighted BP~\citep{WaiJaaWill03b} is optimizing a convexified Bethe energy,
\begin{eqnarray*}
\min_{\{q_{ij}\}_{(i, j\in \Ecal)}} L = \sum_{i}\int_\Hcal q(h_i)\log q(h_i) dh_i + \sum_{i, j}v_{ij}\int_{\Hcal^2} q_{ij}(h_i, h_j)\log\frac{q_{ij}(h_i, h_j)}{q_i(h_i)q_j(h_j)}dh_i dh_j\\
 - \sum_{i}\int_\Hcal q(h_i)\log \Phi(h_i, x_i) dh_i - \sum_{i, j}\int_{\Hcal^2} q_{ij}(h_i, h_j)\log \Psi(h_i, h_j)dh_i dh_j
\end{eqnarray*}
subject to pairwise marginal consistency constraints: $\int_\Hcal q_{ij}(h_i, h_j)dh_j = q_i(h_i)$, $\int_\Hcal q_{ij}(h_i, h_j)dh_j = q_i(h_i) $, and $\int_\Hcal q_i(h_i) dh_i= 1$. The $\{v_{ij}\}_{(i,j)\in \Ecal}$ represents the probabilities that each edge appears in a spanning tree randomly chose from all spanning tree from $G=\{\Vcal, \Ecal\}$ under some measure. Follow the same strategy as loopy BP update derivations, \ie, take derivatives of the corresponding Lagrangian with respect to $q_i$ and $q_{ij}$ and set to zero, meanwhile, incorporate with the marginal consistency, we can arrive the messages updates,
{
\begin{eqnarray*}
m_{ij}(h_j) &\propto& \int_\Hcal \Psi_{ij}^{\frac{1}{v_{ji}}}(h_i, h_j)\Phi_i(h_i, x_i)\frac{\prod_{k\in\Hcal(i)\setminus j}\hspace{-1mm}m_{ki}^{v_{ki}}(h_i)}{m_{ji}^{1-v_{ij}}(h_i)}dh_i,\nonumber\\
q_i(h_i) &\propto& \Phi_i(h_i, x_i)\prod_{k\in \Ncal(i)}m_{ki}^{v_{ki}}(h_i), \nonumber \\
q_{ij}(h_i, h_j)&\propto&\Psi_{ij}^{\frac{1}{v_{ji}}}(h_i, h_j)\Phi_i(h_i, x_i)\Phi_j(h_j, x_j)\frac{\prod_{k\in\Ncal(i)\setminus j}m_{ki}^{v_{ki}}(h_i)}{m_{ji}^{1-v_{ij}}(h_i)}\frac{\prod_{k\in\Ncal(j)\setminus i}m_{kj}^{v_{kj}}(h_j)}{m_{ij}^{1-v_{ji}}(h_j)}.
\end{eqnarray*}
}
Similarly, the embedded messages and the marginals on nodes can be obtained as
\begin{eqnarray*}
\widetilde\nu_{ij} &=& \widetilde\Tcal_1\circ\rbr{x_i, \cbr{\widetilde \nu_{ki}}_{k\in \Ncal(i)\setminus j}, \widetilde \nu_{ji}, \{v_{ki}\}_{k\in \Ncal{i}\setminus j}, v_{ij}},\\
\widetilde\mu_{i} &=& \widetilde\Tcal_2\circ\rbr{x_i, \cbr{\widetilde \nu_{ki}, v_{ki}}_{k\in \Ncal(i)}}.
\end{eqnarray*}
Parametrize these message in the same way, we obtain, 
\begin{eqnarray*}
\widetilde\nu_{ij} &=& \sigma\rbr{W_1x_i + W_2 \sum_{k\in \Ncal(i)\setminus j}\widetilde v_{ki}\nu_{ki} + W_3\widetilde v_{ij}\nu_{ji} },\\
\widetilde\mu_{i} &=& \sigma\rbr{W_4x_i + W_5\sum_{k\in \Ncal(i)}\widetilde v_{ki} \nu_{ki} }.
\end{eqnarray*}
Notice the tree-weighted BP contains extra parameters $\{v_{ij}\}_{(i, j)\in \Ecal}$ which is in the spanning tree polytope as~\citep{WaiJaaWill03b}.

\subsection{Generalized Belief Propagation}

The Kikuchi free energy is the generalization of the Bethe free energy by involving \emph{high-order} interactions. More specifically, given the MRFs, we denote $R$ to be a set of regions, \ie, some basic clusters of nodes, their intersections, the intersections of the intersections, and so on. We denote the $sub(r)$ or $sup(r)$, \ie, subregions or superregions of $r$, as the set of regions completely contained in $r$ or containing $r$, respectively. 
Let $h_r$ be the state of the nodes in region $r$, then, the Kikuchi free energy is 
\begin{eqnarray*}
\sum_{r\in R} c_r\rbr{\int q(h_r)\log\frac{q(h_r)}{\prod_{i, j\in r}\Psi(h_i, h_j)\prod_{i\in r}\Phi(h_i, x_i)}},
\end{eqnarray*}
where $c_r$ is over-counting number of region $r$, defined by $c_r := 1 - \sum_{s\in sup(r)}c_s$ with $c_r = 1$ if $r$ is the largest region in $R$. It is straightforward to verify that the Bethe free energy is a special case of the Kikuchi free energy by setting the basic cluster as pair of nodes. The generalized loopy BP~\citep{YedFreWei01b} is trying to seek the stationary points of the Kikuchi free energy under regional marginal consistency constraints and density validation constraints by following messages updates,
\begin{eqnarray}\label{eq:gbp_messags}
m_{r, s}(h_s)\propto \frac{\int \overline\Psi(h_r, x_{r\setminus s})\mbar_{r\setminus s}(h_{r\setminus s}) dh_{r\setminus s}}{\mbar_{r, s}(h_s)},\nonumber\\
q_r(h_r) \propto \prod_{i\in r}\Phi(h_i, x_i)\prod_{m_{r', s'}\in M(r)}m_{r', s'}(h_{s'}),
\end{eqnarray}
where
\begin{eqnarray*}
\mbar_{r\setminus s}(h_{r\setminus s}) = \prod_{\{r', s'\}\in M(r)\setminus M(s) }m_{r', s'}(h_{s'}),\\[-2mm]
\mbar_{r, s}(h_s) = \prod_{\{r',s'\}\in M(r, s)}m_{r', s'}(h_{s'}),\\[-2mm]
\overline\Psi(h_r, x_{r\setminus s})=\prod_{i,j\in r} \Psi(h_i, h_j)\prod_{i\in r\setminus s}\Phi(h_i, x_i).
\end{eqnarray*}
The $M(r)$ denotes the indices of messages $m_{r', s'}$ that $s'\in sub(r)\cup\{r\}$, and $r'\setminus s'$ is outside $r$. $M(r, s)$ is the set of indices of messages $m_{r', s'}$ where $r'\in sub(r)\setminus s$ and $s'\in sub(s)\cup\{s\}$. 

With the injective embedding assumption for each message $\widetilde\nu_{r, s} = \int \phi(h_s) m_{r, s}(h_s)dh_s$ and $\widetilde\mu_r = \int\phi(h_r)q_r(h_r)dh_r$, following the reasoning~~\eq{eq:embed_funtional} and~\eq{eq:embed_operator}, we can express the embeddings as 
\begin{eqnarray}\label{eq:embed_gbp}
\widetilde\nu_{r, s} &=& \widetilde\Tcal_1\circ\rbr{x_{r\setminus s}, \cbr{\widetilde \nu_{r', s'}}_{M(r)\setminus M(s), M(r, s)}},\\
\widetilde\mu_{r} &=& \widetilde\Tcal_2\circ\rbr{x_r, \cbr{\widetilde \nu_{r', s'}}_{M(r)}}.
\end{eqnarray}
Following the same parameterization in loopy BP, we represent the embeddings by neural network with rectified linear units,
\begin{align}
  \widetilde\nu_{r, s} &=\hspace{-1mm} \sigma\Big(\sum_{i\in r}W^i_1x_r^i + W_2\hspace{-3mm}\sum_{M(r)\setminus M(s)}\hspace{-3mm}\widetilde\nu_{r', s'} + W_3\hspace{-2mm}\sum_{M(r, s)}\hspace{-1mm}\widetilde\nu_{r', s'}\Big)\\[-2mm]
  \widetilde\mu_i &= \sigma\Big(\sum_{i\in r}W_4^ix_i + W_5\sum_{M(r)}\widetilde\nu_{r', s'}\Big)
\end{align}
where $\Wb =\{\{W_1^i\}, W_2, W_3, \{W_4^i\}, W_5\}$ are matrices with appropriate sizes. The generalized BP embedding updates will be almost the same as Algorithm~\ref{alg:lbp_msg} except the order of the iterations. We start from the messages into the smallest region first~\citep{YedFreWei01b}. 

\noindent
{\bf Remark:} The choice of basis clusters and the form of messages determine the dependency in the embedding. Please refer to \citet{YedFreWei05} for details about the principles to partition the graph structure, and several other generalized BP variants with different messages forms. The algorithms proposed for minimizing the Bethe free energy~\citep{Minka01b, Heskes02, Yuille02} can also be extended for Kikuchi free energy, resulting in different embedding forms.

\section{More experiments}\label{appendix:more_exp}

In this section, we present more detailed study of our proposed \textit{structure2vec}. 

\subsection{Graph classification task}

\begin{table*}[t]
\centering
\resizebox{1.0\textwidth}{!}{%
\begin{tabular}{ccccccccccc}
	\toprule
	Datasets & WL subtree & WL edge & WL sp & R\&G & p-rand walk & Rand walk & Graphlet & sp & DE-MF & DE-LBP \\
	\midrule
	MUTAG & 82.05 & 81.06 & 83.78 &  85.72 & 79.19 & 80.72  & 75.61 & 87.28 & 87.78  & 88.28 \\
	\cmidrule{1-11}
	NCI1 & 82.19 & 84.37 & 84.55 & 61.86 & 58.66 & 64.34 & 66.00 & 73.47 & 83.04 & 83.72 \\
	\cmidrule{1-11}
	NCI109 & 82.46 & 84.49 & 83.53 & 61.67 & 58.36 & 63.51 & 66.59 & 73.07 & 82.05 & 82.16\\
	\cmidrule{1-11}
	ENZYMES & 52.22 & 53.17 & 59.05 & 13.35 & 27.67 & 21.68 & 32.70 & 41.68 & 60.17 & 61.10\\
	\cmidrule{1-11}
	DD & 79.78 & 77.95 & 79.43 & 57.27 & 66.64 & 71.70 & 78.59 & 78.45 & 80.68 & 82.22\\
	\bottomrule
\end{tabular}
}
\caption{ 10-fold cross validation accuracy on graph classification benchmark datasets. The 'sp' in the table stands for shortest-path. \label{tab:graph_kernel_results}}
\end{table*}

In addition to the bar plots shown in Figure~\ref{fig:gk_acc}, we present the corresponding raw numerical scores in Table~\ref{tab:graph_kernel_results}.

\section{Derivatives Computation in Algorithm~\ref{alg:framework}}\label{appendix:derivative}

We can use the chain rule to obtain the derivatives with respect to $\Ub^T = \{\Wb^T, \ub^T \}$. According to Equation~\ref{eq:regression} and Equation~\ref{eq:classification}, the message passed to supervised label $y_n$ for $n-$th sample can be represented as $m_y^n = \sum_{i \in \Vcal}\tilde{\mu_i}^n$, and the corresponding derivative can be denoted as 
\[
\frac{\partial l}{\partial m_y^n} = \frac{\partial l}{\partial f} \frac{\partial f}{\partial \sigma(m_y^n) } \frac{\partial \sigma(m_y^n)}{\partial m_y^n}
\]

The term $\frac{\partial l}{\partial f}$ depends on the supervised information and the loss function we used, and $\frac{\partial l}{\partial \sigma(m_y^n) } = \ub^T \frac{\partial l}{\partial f}$. The last term $\frac{\partial \sigma(m_y^n)}{\partial m_y^n}$ depends on the nonlinear function $\sigma$ we used here. 

The derivatives with respect to $\ub$ for the current encountered sample $\{\chi_n, y_n\}$ SGD iteration are
\begin{equation}
  \nabla_{\ub} l(f(\widetilde\mu^n; \Ub), y_n) = \frac{\partial l}{\partial f} \sigma(m_y^n)^T
\end{equation}

In order to update the embedding parameters $\Wb$, we need to obtain the derivatives with respect to the embedding of each hidden node, i.e., $\frac{\partial l}{\partial \tilde{u_i}^n} = \frac{\partial l}{\partial m_y^n}, \forall i \in \Vcal$. 

\textbf{Embedded Mean Field}

In mean field embedding, we unfold the fixed point equation by the iteration index $t = 1, 2, \ldots, T$. At $t-$th iteration, the partial derivative is denoted as $\frac{\partial l}{\partial \tilde{\mu_i}^{n(t)}}$. The partial derivative with respect to the embedding obtained by last round fixed point iteration is already defined above:
$\frac{\partial l}{\partial \tilde{\mu_i}^{n(T)}} = \frac{\partial l}{\partial m_y^n}$

Then the derivatives can be obtained recursively: $\frac{\partial l}{\partial \tilde{\mu_i}^{n(t)}} = \sum_{j, i \in \Ncal(j)} W_2^T \frac{\partial l}{\partial \tilde{\mu_j}^{n(t+1)}} \frac{\partial \sigma}{\partial (W_1x_j + W_2l_j^{(t)}) }$, $t = 1, 2, \ldots, T - 1$. Similarly, the parameters $\Wb$ are also updated cumulatively as below.
\begin{align}
  \frac{\partial l}{\partial (W_1x_i + W_2l_i^{(t)})} = & \sum_{j, i \in \Ncal(j)} \frac{\partial l}{\partial \tilde{\mu_j}^{n(t+1)}} \frac{\partial \sigma}{\partial (W_1x_j + W_2l_j^{(t)}) }  \\
  \nabla_{W_1} l(f(\widetilde\mu^n; \Ub), y_n) = & \sum_{i \in \Vcal_n} \sum_{t=1}^{T-1} \frac{\partial l}{\partial (W_1x_i + W_2l_i^{(t)}) } x_i^T  \\
  \nabla_{W_2} l(f(\widetilde\mu^n; \Ub), y_n) = &  \sum_{i \in \Vcal_n} \sum_{t=1}^{T-1} \frac{\partial l}{\partial (W_1x_i + W_2l_i^{(t)} ) } l_i^{(t)T}
\end{align}

\textbf{Embedding Loopy BP}

Similar as above case, we can first obtain the derivatives with respect to embeddings of hidden variables $\frac{\partial l}{\partial \tilde{\mu_i}^n} = \frac{\partial l}{\partial m_y^n}$. Since the last round of message passing only involves the edge-to-node operations, we can easily get the following derivatives. 
\begin{align}
  \nabla_{W_3} l(f(\widetilde\mu^n; \Ub), y_n) = & \sum_{i \in \Vcal}  \frac{\partial l}{\partial \tilde{\mu_i}^n} \frac{\partial \sigma}{\partial (W_3x_i + W_4\sum_{k \in \Ncal(i)} \widetilde\nu_{ki}^{n(T)})} x_i^T \\
  \nabla_{W_4} l(f(\widetilde\mu^n; \Ub), y_n) = & \sum_{i \in \Vcal} 
  \frac{\partial l}{\partial \tilde{\mu_i}^n} \frac{\partial \sigma}{\partial (W_3x_i + W_4\sum_{k \in \Ncal(i)} \widetilde\nu_{ki}^{n(T)})} (\sum_{k \in \Ncal(i)} \widetilde\nu_{ki}^{n(T)})^T \\
\end{align}

Now we consider the partial derivatives for the pairwise message embeddings for different $t$. Again, the top level one is trivial, which is given by $\frac{\partial l}{\partial \widetilde{\nu_{ij}}^{n(T)} } = W_4^T \frac{\partial l}{\partial \tilde{\mu_j}} \frac{\partial \sigma}{\partial (W_3x_j + W_4\sum_{k \in \Ncal(j)} \widetilde\nu_{kj}^{(T)})} $. Using similar recursion trick, we can get the following chain rule for getting partial derivatives with respect to each pairwise message in each stage of fixed point iteration. 
\begin{align}
  \frac{\partial l}{\partial \widetilde\nu_{ij}^{n(t)}} = & \sum_{p \in \Ncal(j) \setminus i} W_2^T \frac{\partial l}{\partial \widetilde\nu_{jp}^{n(t + 1)}} \frac{\partial \sigma}{\partial (W_1x_j + W_2 \sum_{k \in \Ncal(j) \setminus p}[\widetilde\nu_{kj}^{n(t)}])} 
\end{align}

  Then, we can update the parameters $W_1, W_2$ using following gradients. 
  
\begin{align}
  \nabla_{W_1} l(f(\widetilde\mu^n; \Ub), y_n) = & \sum_{t=1}^{T-1} \sum_{(i, j)\in \Ecal_n} \frac{\partial l}{\partial \widetilde\nu_{ij}^{n(t+1)}} \frac{\partial \sigma}{\partial (W_1x_i + W_2 \sum_{k \in \Ncal(i) \setminus j}[\widetilde\nu_{ki}^{n(t)}])} x_i^T \\
  \nabla_{W_2} l(f(\widetilde\mu^n; \Ub), y_n) = & \sum_{t=1}^{T-1} \sum_{(i, j)\in \Ecal_n} \frac{\partial l}{\partial \widetilde\nu_{ij}^{n(t+1)}} \frac{\partial \sigma}{\partial (W_1x_i + W_2 \sum_{k \in \Ncal(i) \setminus j}[\widetilde\nu_{ki}^{n(t)}])} (\sum_{k \in \Ncal(i) \setminus j}[\widetilde\nu_{ki}^{n(t)}])^T\\
\end{align}

\end{appendix}


\begin{thebibliography}{54}
\providecommand{\natexlab}[1]{#1}
\providecommand{\url}[1]{\texttt{#1}}
\expandafter\ifx\csname urlstyle\endcsname\relax
  \providecommand{\doi}[1]{doi: #1}\else
  \providecommand{\doi}{doi: \begingroup \urlstyle{rm}\Url}\fi

\bibitem[Andreeva et~al.(2004)Andreeva, Howorth, Brenner, Hubbard, Chothia, and
  Murzin]{AndHowBreHubetal04}
Andreeva, A., Howorth, D., Brenner, S.~E., Hubbard, T.~J., Chothia, C., and Murzin, A.~G.
\newblock Scop database in 2004: refinements integrate structure and sequence
  family data.
\newblock \emph{Nucleic acids research}, 32\penalty0 (suppl 1):\penalty0
  D226--D229, 2004.

\bibitem[Borgwardt(2007)]{Borgwardt07}
Borgwardt, K.~M.
\newblock \emph{Graph Kernels}.
\newblock PhD thesis, Ludwig-Maximilians-University, Munich, Germany, 2007.

\bibitem[Borgwardt \& Kriegel(2005)Borgwardt and Kriegel]{BorKri05}
Borgwardt, K.~M. and Kriegel, H.-P.
\newblock Shortest-path kernels on graphs.
\newblock In \emph{ICDM}, 2005.

\bibitem[Bruna et~al.(2013)Bruna, Zaremba, Szlam, and LeCun]{BruZarSzlLeC13}
Bruna, J., Zaremba, W., Szlam, A., and LeCun, Y.
\newblock Spectral networks and locally connected networks on graphs.
\newblock \emph{arXiv preprint arXiv:1312.6203}, 2013.

\bibitem[Chang \& Lin(2001)Chang and Lin]{ChaLin01b}
Chang, C.~C. and Lin, C.~J.
\newblock \emph{{LIBSVM}: a library for support vector machines}, 2001.
\newblock Software available at
  {\small\url{http://www.csie.ntu.edu.tw/~cjlin/libsvm}}.

\bibitem[Chen et~al.(2014)Chen, Schwing, Yuille, and Urtasun]{CheSchYuiUrt14}
Chen, L.~C., Schwing, A.~G., Yuille, A.~L., and Urtasun, R.
\newblock Learning deep structured models.
\newblock \emph{arXiv preprint arXiv:1407.2538}, 2014.

\bibitem[Debnath et~al.(1991)Debnath, Lopez~de Compadre, Debnath, Shusterman,
  and Hansch]{DebLopDebShuetal91}
Debnath, A.~K., Lopez~de Compadre, R.~L., Debnath, G., Shusterman, A.~J., and
  Hansch, C.
\newblock Structure-activity relationship of mutagenic aromatic and
  heteroaromatic nitro compounds. correlation with molecular orbital energies
  and hydrophobicity.
\newblock \emph{J Med Chem}, 34:\penalty0 786--797, 1991.

\bibitem[Dobson \& Doig(2003)Dobson and Doig]{DobDoi03}
Dobson, P.~D. and Doig, A.~J.
\newblock Distinguishing enzyme structures from non-enzymes without alignments.
\newblock \emph{J Mol Biol}, 330\penalty0 (4):\penalty0 771--783, Jul 2003.

\bibitem[Doench et~al.(2014)Doench, Hartenian, Graham, Tothova, Hegde, Smith,
  Sullender, Ebert, Xavier, and Root]{DoeHarGraTotetal14}
Doench, J.~G., Hartenian, E., Graham, D.~B., Tothova, Z., Hegde,
  H., Smith, I., Sullender, M., Ebert, B.~L., Xavier, R.~J.,
  and Root, D.~E.
\newblock Rational design of highly active sgrnas for crispr-cas9-mediated gene
  inactivation.
\newblock \emph{Nature biotechnology}, 32\penalty0 (12):\penalty0 1262--1267,
  2014.

\bibitem[Duvenaud et~al.(2015)Duvenaud, Maclaurin, Iparraguirre, Bombarell,
  Hirzel, Aspuru-Guzik, and Adams]{DuvMacIpaBometal15}
Duvenaud, D.~K., Maclaurin, D., Iparraguirre, J., Bombarell, R.,
  Hirzel, T., Aspuru-Guzik, A., and Adams, R.~P.
\newblock Convolutional networks on graphs for learning molecular fingerprints.
\newblock In \emph{Advances in Neural Information Processing Systems}, pp.\
  2215--2223, 2015.

\bibitem[Fusi et~al.(2015)Fusi, Smith, Doench, and Listgarten]{FusSmiDoeLis15}
Fusi, N., Smith, I., Doench, J., and Listgarten, J.
\newblock In silico predictive modeling of crispr/cas9 guide efficiency.
\newblock \emph{bioRxiv}, 2015.
\newblock \doi{10.1101/021568}.
\newblock URL \url{http://biorxiv.org/content/early/2015/06/26/021568}.

\bibitem[G\"artner et~al.(2003)G\"artner, Flach, and Wrobel]{GaeFlaWro03}
G\"artner, T., Flach, P.A., and Wrobel, S.
\newblock On graph kernels: Hardness results and efficient alternatives.
\newblock In Sch\"olkopf, B. and Warmuth, M.~K. (eds.), \emph{Proceedings of Annual Conference.\ Computational Learning Theory}, pp.\  129--143. Springer, 2003.

\bibitem[Giles(2001)]{CarLawGil01}
Caruana, R., Lawrence, S., and Giles, L.
\newblock Overfitting in neural nets: Backpropagation, conjugate gradient, and
  early stopping.
\newblock In \emph{Advances in Neural Information Processing Systems 13}, volume~13, pp.\  402. MIT Press, 2001.

\bibitem[Hachmann et~al.(2011)Hachmann, Olivares-Amaya, Atahan-Evrenk,
  Amador-Bedolla, S{\'a}nchez-Carrera, Gold-Parker, Vogt, Brockway, and
  Aspuru-Guzik]{HacOliAtaAmaetal11}
Hachmann, J., Olivares-Amaya, R., Atahan-Evrenk, S.,
  Amador-Bedolla, C., S{\'a}nchez-Carrera, R.~S., Gold-Parker, A.,
  Vogt, L., Brockway, A.~M., and Aspuru-Guzik, A.
\newblock The harvard clean energy project: large-scale computational screening
  and design of organic photovoltaics on the world community grid.
\newblock \emph{The Journal of Physical Chemistry Letters}, 2\penalty0
  (17):\penalty0 2241--2251, 2011.

\bibitem[Henaff et~al.(2015)Henaff, Bruna, and LeCun]{HenBruLeC15}
Henaff, M., Bruna, J., and LeCun, Y.
\newblock Deep convolutional networks on graph-structured data.
\newblock \emph{arXiv preprint arXiv:1506.05163}, 2015.

\bibitem[Hershey et~al.(2014)Hershey, Roux, and Weninger]{HerRouWen14}
Hershey, J.~R., Roux, J.~L., and Weninger, F.
\newblock Deep unfolding: Model-based inspiration of novel deep architectures.
\newblock \emph{arXiv preprint arXiv:1409.2574}, 2014.

\bibitem[Heskes(2002)]{Heskes02}
Heskes, T.
\newblock Stable fixed points of loopy belief propagation are local minima of
  the bethe free energy.
  \emph{Advances in Neural Information Processing Systems}, pp.\  343--350. MIT Press, 2002.

\bibitem[Jaakkola \& Haussler(1999)Jaakkola and Haussler]{JaaHau99b}
Jaakkola, T.~S. and Haussler, D.
\newblock Exploiting generative models in discriminative classifiers.
\newblock In Kearns, M.~S., Solla, S.~A., and Cohn, D.~A. (eds.),
  \emph{Advances in Neural Information Processing Systems 11}, pp.\  487--493.
  {MIT} Press, 1999.

\bibitem[Jebara et~al.(2004)Jebara, Kondor, and Howard]{JebKonHow04}
Jebara, T., Kondor, R., and Howard, A.
\newblock Probability product kernels.
\newblock \emph{J. Mach. Learn. Res.}, 5:\penalty0 819--844, 2004.

\bibitem[Jitkrittum et~al.(2015)Jitkrittum, Gretton, Heess, Eslami,
  Lakshminarayanan, Sejdinovic, and Szab{\'{o}}]{JitGreHeeEsletal15}
Jitkrittum, W., Gretton, A., Heess, N., Eslami, S. M.~A.,
  Lakshminarayanan, B., Sejdinovic, D., and Szab{\'{o}}, Z.
\newblock Kernel-based just-in-time learning for passing expectation
  propagation messages.
\newblock In \emph{Proceedings of the Thirty-First Conference on Uncertainty in
  Artificial Intelligence, {UAI} 2015, July 12-16, 2015, Amsterdam, The
  Netherlands}, pp.\  405--414, 2015.

\bibitem[Kuang et~al.(2005)Kuang, Ie, Wang, Wang, Siddiqi, Freund, and
  Leslie]{KuaIeWanWanetal05}
Kuang, R., Ie, E., Wang, K., Wang, K., Siddiqi, M., Freund, Y., and
  Leslie, C.
\newblock Profile-based string kernels for remote homology detection and motif
  extraction.
\newblock \emph{Journal of bioinformatics and computational biology},
  3\penalty0 (03):\penalty0 527--550, 2005.

\bibitem[Landrum(2012)]{Landrum12}
Landrum, G.
\newblock Rdkit: Open-source cheminformatics (2013), 2012.

\bibitem[Leslie et~al.(2002{\natexlab{a}})Leslie, Eskin, and
  Noble]{LesEskNob02}
Leslie, C., Eskin, E., and Noble, W.~S.
\newblock The spectrum kernel: A string kernel for {SVM} protein
  classification.
\newblock In \emph{Proceedings of the Pacific Symposium on Biocomputing}, pp.\
  564--575, Singapore, 2002{\natexlab{a}}. World Scientific Publishing.

\bibitem[Leslie et~al.(2002{\natexlab{b}})Leslie, Eskin, Weston, and
  Noble]{LesEskWesNob02}
Leslie, C., Eskin, E., Weston, J., and Noble, W.~S.
\newblock Mismatch string kernels for {SVM} protein classification.
\newblock In \emph{Advances in
  Neural Information Processing Systems}, volume~15, Cambridge, MA,
  2002{\natexlab{b}}. {MIT} Press.

\bibitem[Li et~al.(2015)Li, Tarlow, Brockschmidt, and Zemel]{LiTarBroZem15}
Li, Y., Tarlow, D., Brockschmidt, M., and Zemel, R..
\newblock Gated graph sequence neural networks.
\newblock \emph{arXiv preprint arXiv:1511.05493}, 2015.

\bibitem[Lin et~al.(2015)Lin, Shen, Reid, and {van den Hengel}]{LinSheReiHen15}
Lin, G., Shen, C., Reid, I., and {van den Hengel}, A.
\newblock Deeply learning the messages in message passing inference.
\newblock In \emph{Advances in Neural Information Processing Systems}, 2015.

\bibitem[Minka(2001)]{Minka01b}
Minka, T.
\newblock {The EP energy function and minimization schemes}.
\newblock \emph{See www. stat. cmu. edu/minka/papers/learning. html, August},
  2001.

\bibitem[Mou et~al.(2016)Mou, Li, Zhang, Wang, and Jin]{MouLiZhaWanetal16}
Mou, L., Li, G., Zhang, L., Wang, T., and Jin, Z..
\newblock Convolutional neural networks over tree structures for programming
  language processing.
\newblock In \emph{Proceedings of the Thirtieth AAAI Conference on Artificial
  Intelligence}, 2016.

\bibitem[Murphy et~al.(1999)Murphy, Weiss, and Jordan]{MurWeiJor99}
Murphy, K.~P., Weiss, Y., and Jordan, M.~I.
\newblock Loopy belief propagation for approximate inference: An empirical
  study.
\newblock In \emph{UAI}, pp.\  467--475, 1999.

\bibitem[Pearl(1988)]{Pearl88}
Pearl, J.
\newblock \emph{Probabilistic Reasoning in Intelligent Systems: Networks of
  Plausible Inference}.
\newblock Morgan Kaufman, 1988.

\bibitem[Pyzer-Knapp et~al.(2015)Pyzer-Knapp, Li, and Aspuru-Guzik]{PyzLiAsp15}
Pyzer-Knapp, E.~O., Li, K., and Aspuru-Guzik, A.
\newblock Learning from the harvard clean energy project: The use of neural
  networks to accelerate materials discovery.
\newblock \emph{Advanced Functional Materials}, 25\penalty0 (41):\penalty0
  6495--6502, 2015.

\bibitem[Ramon \& G\"{a}rtner(2003)Ramon and G\"{a}rtner]{RamGae03}
Ramon, J. and G\"{a}rtner, T.
\newblock Expressivity versus efficiency of graph kernels.
\newblock Technical report, First International Workshop on Mining Graphs,
  Trees and Sequences (held with ECML/PKDD'03), 2003.

\bibitem[Ross et~al.(2011)Ross, Munoz, Hebert, and Bagnell]{RosMunHebBag11}
Ross, S., Munoz, D., Hebert, M., and Bagnell, J.~A.
\newblock Learning message-passing inference machines for structured
  prediction.
\newblock In \emph{IEEE
  Conference on Computer Vision and Pattern Recognition}, pp.\  2737--2744. IEEE, 2011.

\bibitem[Scarselli et~al.(2009)Scarselli, Gori, Tsoi, Hagenbuchner, and
  Monfardini]{ScaGorTsoHagetal09}
Scarselli, F., Gori, M., Tsoi, A.~C., Hagenbuchner, M., and
  Monfardini, G.
\newblock The graph neural network model.
\newblock \emph{IEEE Transactions on Neural Networks}, 20\penalty0
  (1):\penalty0 61--80, 2009.

\bibitem[Sch{\"o}lkopf et~al.(2004)Sch{\"o}lkopf, Tsuda, and Vert]{SchTsuVer04}
Sch{\"o}lkopf, B., Tsuda, K., and Vert, J.-P.
\newblock \emph{Kernel Methods in Computational Biology}.
\newblock MIT Press, Cambridge, MA, 2004.

\bibitem[Sch{\"o}lkopf \& Smola(2002)Sch{\"o}lkopf and Smola]{SchSmo02}
Sch{\"o}lkopf, B., and Smola, A.~J.
\newblock \emph{Learning with Kernels}.
\newblock {MIT} Press, Cambridge, MA, 2002.

\bibitem[Shervashidze et~al.(2009)Shervashidze, Vishwanathan, Petri, Mehlhorn,
  and Borgwardt]{SheVisPetMehetal09}
Shervashidze, N., Vishwanathan, S.~V.~N., Petri, T., Mehlhorn, K., and
  Borgwardt, K.
\newblock Efficient graphlet kernels for large graph comparison.
\newblock \emph{Proceedings of International Conference on Artificial Intelligence and Statistics}. Society for Artificial
  Intelligence and Statistics, 2009.

\bibitem[Shervashidze et~al.(2011)Shervashidze, Schweitzer, Van~Leeuwen,
  Mehlhorn, and Borgwardt]{SheSchVanMehetal11}
Shervashidze, N., Schweitzer, P., Van~Leeuwen, E.~J., Mehlhorn, K.,
  and Borgwardt, K.~M.
\newblock Weisfeiler-lehman graph kernels.
\newblock \emph{The Journal of Machine Learning Research}, 12:\penalty0
  2539--2561, 2011.

\bibitem[Smola et~al.(2007)Smola, Gretton, Song, and
  {Sch\"olkopf}]{SmoGreSonSch07}
Smola, A.~J., Gretton, A., Song, L., and {Sch\"olkopf}, B.
\newblock A {H}ilbert space embedding for distributions.
\newblock In \emph{Proceedings of the International Conference on Algorithmic
  Learning Theory}, volume 4754, pp.\  13--31. Springer, 2007.

\bibitem[Song et~al.(2009)Song, Huang, Smola, and Fukumizu]{SonHuaSmoFuk09}
Song, L., Huang, J., Smola, A.~J., and Fukumizu, K.
\newblock Hilbert space embeddings of conditional distributions.
\newblock In \emph{Proceedings of the International Conference on Machine
  Learning}, 2009.

\bibitem[Song et~al.(2010)Song, Gretton, and Guestrin]{SonGreGue10}
Song, L., Gretton, A., and Guestrin, C.
\newblock Nonparametric tree graphical models.
\newblock In \emph{13th Workshop on Artificial Intelligence and Statistics},
  volume~9 of \emph{JMLR workshop and conference proceedings}, pp.\  765--772,
  2010.

\bibitem[Song et~al.(2011)Song, Gretton, Bickson, Low, and
  Guestrin]{SonGreBicLowGue11}
Song, L., Gretton, A., Bickson, D., Low, Y., and Guestrin, C.
\newblock Kernel belief propagation.
\newblock In \emph{Proc.\ Intl.\ Conference on Artificial Intelligence and
  Statistics}, volume~10 of \emph{JMLR workshop and conference proceedings},
  2011.

\bibitem[Sriperumbudur et~al.(2008)Sriperumbudur, Gretton, Fukumizu, Lanckriet,
  and Sch{\"{o}}lkopf]{SriGreFukLanetal08}
Sriperumbudur, B., Gretton, A., Fukumizu, K., Lanckriet, G., and
  Sch{\"{o}}lkopf, B.
\newblock Injective {H}ilbert space embeddings of probability measures.
\newblock In \emph{Proceedings of Annual Conference.\ Computational Learning Theory}, pp.\
  111--122, 2008.

\bibitem[Sugiyama \& Borgwardt(2015)Sugiyama and Borgwardt]{SugBor15}
Sugiyama, M. and Borgwardt, K.
\newblock Halting in random walk kernels.
\newblock In \emph{Advances in Neural Information Processing Systems}, pp.\
  1630--1638, 2015.

\bibitem[Vishwanathan \& Smola(2003)Vishwanathan and Smola]{VisSmo03}
Vishwanathan, S.~V.~N. and Smola, A.~J.
\newblock Fast kernels for string and tree matching.
\newblock In Becker, S., Thrun, S., and Obermayer, K. (eds.), \emph{Advances in
  Neural Information Processing Systems 15}, pp.\  569--576. {MIT} Press,
  Cambridge, MA, 2003.

\bibitem[Vishwanathan et~al.(2010)Vishwanathan, Schraudolph, Kondor, and
  Borgwardt]{VisSchKonBor10}
Vishwanathan, S.~V.~N., Schraudolph, N.~N., Kondor, I.~R., and
  Borgwardt, K.~M.
\newblock Graph kernels.
\newblock \emph{Journal of Machine Learning Research}, 2010.
\newblock URL
  \url{http://www.stat.purdue.edu/~vishy/papers/VisSchKonBor10.pdf}.
\newblock In press.

\bibitem[Wainwright et~al.(2003)Wainwright, Jaakkola, and
  Willsky]{WaiJaaWill03b}
Wainwright, M., Jaakkola, T., and Willsky, A.
\newblock Tree-reweighted belief propagation and approximate {ML} estimation by
  pseudo-moment matching.
\newblock In \emph{9th Workshop on Artificial Intelligence and Statistics},
  2003.

\bibitem[Wainwright \& Jordan(2008)Wainwright and Jordan]{WaiJor08}
Wainwright, M.~J. and Jordan, M.~I.
\newblock Graphical models, exponential families, and variational inference.
\newblock \emph{Foundations and Trends in Machine Learning}, 1\penalty0 (1 --
  2):\penalty0 1--305, 2008.

\bibitem[Wale et~al.(2008)Wale, Watson, and Karypis]{WalWatKar08}
Wale, N., Watson, I.~A., and Karypis, G.
\newblock Comparison of descriptor spaces for chemical compound retrieval and
  classification.
\newblock \emph{Knowledge and Information Systems}, 14\penalty0 (3):\penalty0
  347--375, 2008.

\bibitem[Yedidia et~al.(2001{\natexlab{a}})Yedidia, Freeman, and
  Weiss]{YedFreWei01}
Yedidia, J.~S., Freeman, W.~T., and Weiss, Y.
\newblock Generalized belief propagation.
\newblock \emph{Advances in Neural Information Processing Systems}, pp.\  689--695.
  {MIT} Press, 2001{\natexlab{a}}.

\bibitem[Yedidia et~al.(2001{\natexlab{b}})Yedidia, Freeman, and
  Weiss]{YedFreWei01b}
Yedidia, J.S., Freeman, W.T., and Weiss, Y.
\newblock Bethe free energy, kikuchi approximations and belief propagation
  algorithms.
\newblock Technical report, Mitsubishi Electric Research Laboratories,
  2001{\natexlab{b}}.

\bibitem[Yedidia et~al.(2005)Yedidia, Freeman, and Weiss]{YedFreWei05}
Yedidia, J.S., Freeman, W.T., and Weiss, Y.
\newblock Constructing free-energy approximations and generalized belief
  propagation algorithms.
\newblock \emph{IEEE Transactions on Information Theory}, 51\penalty0
  (7):\penalty0 2282--2312, 2005.

\bibitem[Yuille(2002)]{Yuille02}
Yuille, A.~L.
\newblock Cccp algorithms to minimize the bethe and kikuchi free energies:
  Convergent alternatives to belief propagation.
\newblock \emph{Neural Computation}, 14:\penalty0 2002, 2002.

\bibitem[Zheng et~al.(2015)Zheng, Jayasumana, Romera-Paredes, Vineet, Su, Du,
  Huang, and Torr]{ZheJayRomVinetal15}
Zheng, S., Jayasumana, S., Romera-Paredes, B., Vineet, B.,
  Su, Z., Du, D., Huang, C., and Torr, P.
\newblock Conditional random fields as recurrent neural networks.
\newblock \emph{arXiv preprint arXiv:1502.03240}, 2015.

\end{thebibliography}
\end{document}